\documentclass[11pt]{article}

\usepackage[preprint]{acl}

\usepackage{times}
\usepackage{latexsym}

\usepackage[T1]{fontenc}
\usepackage[utf8]{inputenc}

\usepackage{microtype}

\usepackage{inconsolata}

\usepackage{graphicx}

\usepackage{colortbl} 
\usepackage{hyperref}
\hypersetup{
  colorlinks=true,
  linkcolor=orange!50!brown,
  citecolor=teal,
  urlcolor=magenta,
  filecolor=magenta,
}
\usepackage{url}
\usepackage{array}
\usepackage{graphicx}
\usepackage{amsmath,amssymb}
\usepackage{algorithm}
\usepackage{algpseudocode}
\newcommand{\fs}{\cellcolor[gray]{0.95}}
\usepackage{booktabs}
\title{ReLaDrift: Recurrent Language Generation via Drifting for Test-Time Scaling}

\author{
  \textbf{Yuki Otsuka\textsuperscript{1}},
  \textbf{Daisuke Oba\textsuperscript{1}},
  \textbf{Naoaki Okazaki\textsuperscript{1,2,3}}
\\
  \textsuperscript{1}Institute of Science Tokyo,
  \textsuperscript{2}AIST,
  \textsuperscript{3}NII LLMC
\\
  \texttt{\{yuki.otsuka@nlp., daisuke.oba@nlp., okazaki@\}comp.isct.ac.jp}
}

\begin{document}
\maketitle
\begin{abstract}
Drifting Models generate a sample with a single function evaluation, 
making them attractive for fast parallel language generation, 
but their one-step formulation does not support iterative refinement.
We propose Recurrent Language Generation via Drifting (ReLaDrift), which uses a shared Drifting Model to generate an initial draft from noise and iteratively refine it in a continuous latent space.
We train this recurrence by sampling the number of rounds and applying the drifting loss only at the final round, stopping gradients through preceding rounds so that only one round requires backpropagation.
Unlike the one-step Drifting Model, whose accuracy does not improve when iterated, ReLaDrift enables a trade-off between test-time computation and generation quality, as diffusion and flow language models do.
On Sudoku Hard and GSM8K, it achieves competitive or superior performance compared with these iterative-generation approaches across a range of inference budgets.
Our code is available at \href{https://github.com/Yuki-Ots/ReLaDrift}{\texttt{github.com/Yuki-Ots/ReLaDrift}}.
\end{abstract}

\begin{figure}[!t]
\centering
\includegraphics[width=0.9\columnwidth]{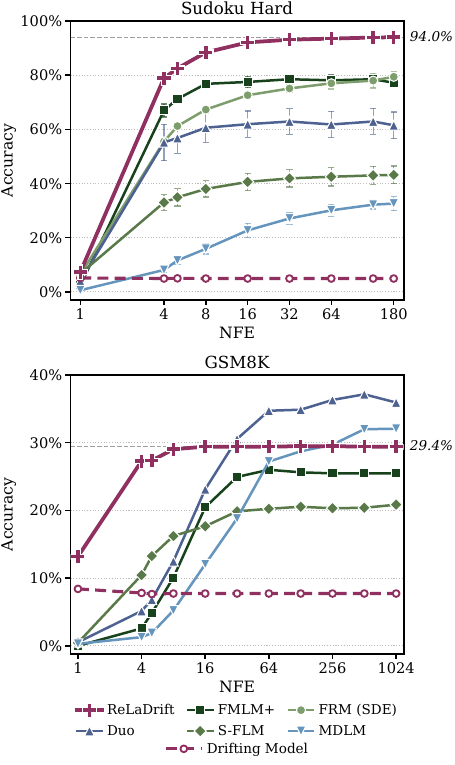}
\caption{
\textbf{Accuracy vs. the number of function evaluations (NFE) on Sudoku Hard (top) and GSM8K (bottom).}
Results use a single generation trajectory and EMA weights, with mean $\pm$ standard deviation over three seeds for Sudoku.
Models are evaluated at 20,000 updates (Sudoku) and 250,000 updates (GSM8K); sampling settings are detailed in Tables~\ref{tab:sudoku_results} and~\ref{tab:gsm8k_results}.}
\label{fig:teaser}
\end{figure}

\section{Introduction}\label{sec:introduction}

The Drifting Model~\citep{deng2026generativemodelingdrifting} learns a direct mapping from noise to data using a drift field computed from real and generated samples during training, enabling generation in a single function evaluation at inference time.
While attractive for fast language generation, this one-step formulation offers no mechanism to improve outputs with additional test-time computation, a form of test-time scaling~\citep{snell2025scaling}.
In contrast, diffusion and flow language models can trade additional sampling steps for improved generation quality~\citep{austin2021d3pm,sahoo2024mdlm,sahoo2025duo,nie2025llada,li2022diffusionlm,lee2026fmlm,deschenaux2026sflm,hu2026elf}.

Simply iterating the Drifting Model does not guarantee improvement,
as it is trained to generate from noise rather than refine its own outputs.
Indeed, Figure~\ref{fig:teaser} shows that simply iterating a one-step Drifting Model yields no accuracy gains on either Sudoku Hard~\citep{alp2024sudoku} or GSM8K~\citep{cobbe2021gsm8k}.
These observations raise the central question of this work:
\emph{how can we train a one-step Drifting Model to refine its own outputs, enabling a trade-off between inference-time computation and generation quality?}

We propose \textbf{Recurrent Language Generation via Drifting (ReLaDrift)}, which uses a shared Drifting Model to generate an initial draft from noise and iteratively refine it in a continuous latent space over successive rounds.
During training, we sample the number of rounds and repeatedly apply the shared model, starting from noise and feeding each draft into the next round.
We apply the drifting loss only to the final output, stopping gradients through the preceding rounds to avoid backpropagation through time (Sec.~\ref{sec:method}).

We evaluate ReLaDrift on Sudoku Hard~\citep{alp2024sudoku} and GSM8K~\citep{cobbe2021gsm8k}, two reasoning tasks with verifiable answers that allow us to measure test-time scaling directly. The tasks present complementary challenges: Sudoku requires a unique solution satisfying global constraints, whereas GSM8K admits multiple valid Python programs~\citep{gao2023pal}.
Our results demonstrate that, by using ReLaDrift, drifting-based generation can trade additional test-time computation for improved accuracy,
achieving competitive or superior performance to established inference-time refinement approaches, including diffusion and flow language models, 
across a range of inference budgets~(Sec.~\ref{sec:experiments}).

\section{Related Work}
\paragraph{Iterative generation vs. one-step drifting.}
Diffusion and flow language models iteratively refine samples at inference
time, allowing a trade-off between generation quality and computational
cost~\citep{austin2021d3pm,sahoo2024mdlm,sahoo2025duo,nie2025llada,li2022diffusionlm,lee2026fmlm,deschenaux2026sflm,hu2026elf}. 
Improving this trade-off has been an active area of research~\citep{
deschenaux2025sdtt,lee2026fmlm,song2023consistency,
frans2025shortcut,geng2025meanflow,boffi2025flowmap,oba2026stopping}.
In contrast, the Drifting Model~\citep{deng2026generativemodelingdrifting}
is designed for one-step generation, with no built-in mechanism
to benefit from additional inference steps, as it is trained
to generate from noise rather than refine its own outputs.

\paragraph{Drifting-based methods.}
Drift Flow Matching~\citep{ma2026dfm} bridges drifting and flow matching by learning transport between arbitrary time points along a noise-to-data interpolation, enabling test-time scaling through finer transport steps, as demonstrated in image generation and robotic control.
\citet{oba2026tokendrift} apply the drifting objective to improve diffusion language models, rather than using the drifting map itself as a generator.
In contrast, ReLaDrift first generates a complete draft in one step, as in the original Drifting Model, and then refines it recurrently without relying on a predefined noise-to-data trajectory.

\paragraph{Recurrent refinement.}
Iterative refinement has been widely studied in parallel language
generation and diffusion models~\citep{lee2018iterative,
ghazvininejad2019maskpredict,gu2019levenshtein,
savinov2022sundae,wang2025remdm,agarwal2026posterior}.
Recurrent refinement has also been explored in flow-based
generation~\citep{helbling2026frm}.
More broadly, weight-tied recurrent models leverage repeated
compute to improve predictions~\citep{
dehghani2019universal,bai2019deq,geiping2025recurrent}.

\section{Background: Drifting Model}\label{sec:drifting}

The Drifting Model~\citep{deng2026generativemodelingdrifting}
learns a direct mapping $f_\theta$ from noise
$\epsilon \sim p_{\mathrm{init}}$ to generated samples $x=f_\theta(\epsilon)$.
The resulting distribution $q_\theta=(f_\theta)_\#p_{\mathrm{init}}$
is trained to match the target data distribution $p$.
Like GANs~\citep{goodfellow2014gan},
it generates samples in a single step, but learns without adversarial training.
Instead, during training, a drift field computed from real
and generated samples guides $q_\theta$ toward the target
distribution $p$.
The field combines attraction toward real samples
with repulsion from generated samples:
\[
\begin{aligned}
V_{p,q_\theta}(x) &= V_p^+(x) - V_{q_\theta}^-(x),\\
V_p^+(x) &=
\frac{\mathbb{E}_{y\sim p}[k(x,y)(y-x)]}
{\mathbb{E}_{y\sim p}[k(x,y)]},\\
V_{q_\theta}^-(x) &=
\frac{\mathbb{E}_{y\sim q_\theta}[k(x,y)(y-x)]}
{\mathbb{E}_{y\sim q_\theta}[k(x,y)]}.
\end{aligned}
\]
where $k(x,y)=\exp(-\|x-y\|/\tau)$ is a kernel with bandwidth $\tau$.

In practice, the expectations are estimated from mini-batches
of real (\emph{positive}) and generated (\emph{negative}) samples.
The generator learns to move its outputs along the drift field
through a stop-gradient regression loss
\(
\mathcal{L}_{\mathrm{drift}}(\theta)
=
\mathbb{E}_{\epsilon \sim p_{\mathrm{init}}}
\left\|
f_\theta(\epsilon) -
\operatorname{sg}\left(
f_\theta(\epsilon) + V_{p,q_\theta}(f_\theta(\epsilon))
\right)
\right\|^2,
\)
where $\operatorname{sg}$ denotes stop-gradient.
The drift field guides training but is not needed at inference time.
We follow the original model's distance and field normalization
and multi-temperature kernels, with implementation-specific
adjustments detailed in Appendix~\ref{app:drift_field}.

\begin{figure*}[t]
\centering
\includegraphics[width=\textwidth]{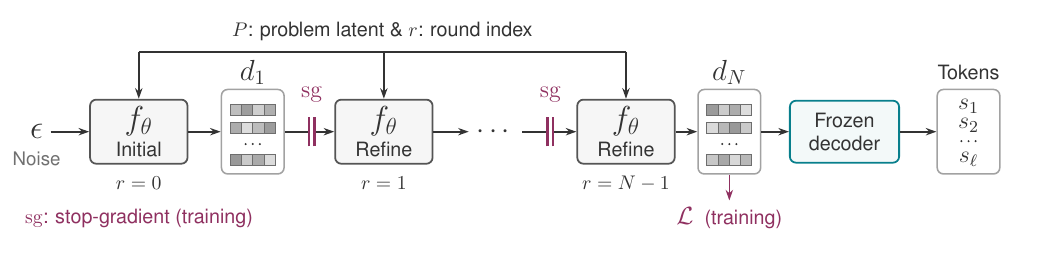}
\caption{\textbf{ReLaDrift in the source formulation.} Round 0 (Initial) transports the noise \(\epsilon\) to the first draft \(d_1\) through the input projection \(W_{\mathrm{noise}}\), and every later round (Refine) transports the current draft through \(W_{\mathrm{draft}}\); all rounds share the network \(f_\theta\) and receive the prompt latent \(P\) and the round index \(r\) (Sec.~\ref{sec:recurrence}). The frozen decoder is applied once to the last draft \(d_N\); at test time, the number of rounds \(N\) is the NFE and may exceed the training depth \(R_{\max}\) (Sec.~\ref{sec:generation}). The marks \(\operatorname{sg}\) and the loss \(\mathcal{L}\) apply only during training, where \(N=R\sim\operatorname{Uniform}\{1,\ldots,R_{\max}\}\) and the drafts that enter a round are treated as constants, so the loss backpropagates through the terminal round only (Sec.~\ref{sec:training_additive}). The positives and the negatives of the loss are not shown.}
\label{fig:reladrift_overview}
\end{figure*}

\section{Methods}\label{sec:method}

ReLaDrift applies one Drifting Model recurrently: the first round transports noise to a first draft, as a standard Drifting Model does, and each later round feeds the current draft back into the same model and returns a corrected draft (Fig.~\ref{fig:reladrift_overview}).
The number of rounds at test time can then be chosen freely.
In the following, we consider conditional generation where a problem statement is given.

\subsection{Continuous Representation of Token Sequences}\label{sec:embeddings}

Language is discrete, so we first embed each token sequence in a continuous space where the Drifting Model can operate.
A fixed encoder maps a token sequence \(s=[s_1,\ldots,s_\ell]\) to one
\(D_{\mathrm{emb}}\)-dimensional vector per position,
\(\operatorname{Encoder}(s)\in\mathbb{R}^{\ell\times D_{\mathrm{emb}}}\).
The encoder is not trained with the model, as in latent generation with frozen pretrained encoders~\citep{lovelace2023ld4lg,shabalin2025tencdm,hu2026elf}, and any map from token sequences to a continuous space can be used in principle; the choice matters (Section~\ref{sec:gsm8k}).

A solution in the training data is represented by a \emph{positive} \(y\in\mathbb{R}^{L\times D_{\mathrm{emb}}}\), the encoding of the token sequence at the \(L\) positions that the model outputs (Section~\ref{sec:experiments}; Appendix~\ref{app:encoding}).
Only the positions of the solution are generated: the noise, the draft, the drift field, and the loss are restricted to these \emph{generated positions}.

The output of the model is continuous, so a decoder maps it back to tokens.
The decoder is either given by the encoder or trained separately in advance, and is frozen while \(f_\theta\) is trained (Appendix~\ref{app:decoder}).
It predicts every token from the latent sequence in parallel, so decoding adds
one pass and does not reintroduce sequential generation.

\subsection{Recurrent Application of the Drifting Model}\label{sec:recurrence}

ReLaDrift applies the network \(f_\theta\) of the Drifting Model
(Section~\ref{sec:drifting}) repeatedly: each application is a \emph{round},
and its output is a \emph{draft}.
Round \(r=0\) produces the first draft
\(d_1\in\mathbb{R}^{L\times D_{\mathrm{emb}}}\) from noise, and each later
round \(r\geq1\) takes the previous draft \(d_r\) and produces the next draft
\(d_{r+1}\).
Thus \(N\) rounds are \(r=0,\ldots,N-1\), and round \(r\) returns \(d_{r+1}\).
Every round also receives the prompt latent \(P\) and the round index \(r\).
All rounds use the same network: the round index selects the input and
output projections of the first round or of the refinement rounds (see
below), so the refinement rounds \(r\geq1\) share all their parameters.
Drafts are passed between rounds as continuous latents; they are not decoded
into tokens and re-encoded (Figure~\ref{fig:reladrift_overview}).

The noise and the draft enter \(f_\theta\) through separate input projections,
and we consider three ways to feed the draft back.
In the source formulation, a round has a single input: the first round
transports noise, and each later round transports the previous draft, which
replaces the noise as the input of the round but keeps its own projection.
In the additive formulation, every round receives fresh noise and the draft,
whose projections are added; in the conditional formulation, every round
receives fresh noise, and the draft is appended to the input sequence as extra
tokens.
The additive and conditional formulations therefore need a fixed initial
draft \(d_0\) at the first round, whereas the source formulation does not.
The source formulation is used in all main experiments, and the three
formulations are compared on Sudoku (Section~\ref{sec:sudoku}).

\paragraph{Source formulation.}\label{sec:source_formulation}
A round has a single input \(x\), the noise at \(r=0\) and the previous draft at \(r\geq1\), written \(f_\theta(x,P,r)\).
Inside \(f_\theta\), the projection selected by the round index, \(W_{\mathrm{noise}}\) at \(r=0\) and \(W_{\mathrm{draft}}\) at \(r\geq1\), maps the input to the model dimension, giving the input \(u_r\) of the bidirectional Transformer at the \(L\) generated positions:
\[
\begin{aligned}
d_1&=f_\theta(\epsilon,P,0), & u_0&=\epsilon W_{\mathrm{noise}},\\
d_{r+1}&=f_\theta(d_r,P,r), & u_r&=d_r W_{\mathrm{draft}},
\end{aligned}
\]
for \(r\geq1\), where \(\epsilon\sim p_{\mathrm{init}}=\mathcal{N}(0,I)\) and
\(W_{\mathrm{noise}},W_{\mathrm{draft}}\in\mathbb{R}^{D_{\mathrm{emb}}\times D_{\mathrm{model}}}\).
The noise and the draft thus never share a projection, and the projection that
a round does not select has no effect on that round, so the gradient of that round's loss with respect to it is zero.
For a fixed prompt and round \(r\geq1\), \(x\mapsto f_\theta(x,P,r)\) maps
\(\mathbb{R}^{L\times D_{\mathrm{emb}}}\) to itself, which allows the output of one round to be the input of the next.
No new noise is drawn after the first round, so the refinement path at test time is deterministic once the first draft is fixed.

\paragraph{Additive and conditional formulations.}\label{sec:additive}\label{sec:conditional_formulation}
Both keep the original form of the Drifting Model, which transports fresh noise
at every round, and condition it on the previous draft, so a round has two
inputs: \(d_{r+1}=f_\theta(\epsilon_r,d_r,P,r)\) with
\(\epsilon_r\sim p_{\mathrm{init}}\) for \(r\geq0\), starting from a fixed
initial draft \(d_0\) (Appendix~\ref{app:architecture}); the refinement path
thus remains stochastic.
In the additive formulation, the projected draft is added to the projected
noise at every generated position,
\(u_r=\epsilon_r W_{\mathrm{noise}}+d_r W_{\mathrm{draft}}\), so the draft
enters the model at the same positions as the noise.
In the conditional formulation, the projected draft is instead appended to the
input sequence as \(L\) separate tokens, so that
\(u_r=\epsilon_r W_{\mathrm{noise}}\) and every position of the current
generation can attend to every position of the draft without mixing the two,
at the cost of a longer attention sequence.

\paragraph{Shared components.}
The three formulations differ only in how the noise and the draft enter the Transformer; the rest is shared.
In every formulation, the projected prompt latent \(P W_{\mathrm{prompt}}\) and
the projected round embedding \(\phi(r)W_{\mathrm{round}}\) are also given to
the Transformer, and its output at the generated positions is projected
linearly back to \(D_{\mathrm{emb}}\), with different output projections for
the first round (\(r=0\)) and the refinement rounds (\(r\geq1\)).
The round embedding \(\phi(r)\), a fixed vector of features of \(r\) without
learned parameters, lets
the model distinguish the rounds, whose input distributions differ (in the source formulation, Gaussian noise at the first round and model-produced drafts later).
Appendix~\ref{app:architecture} gives the exact form of all these components, including the placement of the prompt latent in the input sequence for each task.

\subsection{Training}\label{sec:training_additive}

\paragraph{Loss at one round.}
Each round is trained with the loss of the Drifting Model
(Section~\ref{sec:drifting}), with the positives of a problem as samples of the
target distribution \(p\) and the drafts that the round generates for the
problem as samples of the generated distribution \(q\).
For the positives \(y_1,\ldots,y_{N_{+}}\) and the \(n\) drafts
\(x_1,\ldots,x_n\) of a problem, which we call its \emph{generated samples}, the
field and the loss are
\[
\begin{aligned}
V_j&=V^{+}_j-V^{-}_j,\\
\mathcal{L}(\theta)
&=
\frac{1}{n}
\sum_{j=1}^{n}
\bigl\|
x_j-\operatorname{sg}\bigl(x_j+V_j\bigr)
\bigr\|_G^2 ,
\end{aligned}
\]
where \(\operatorname{sg}\) denotes the stop-gradient operation and \(\|\cdot\|_G\) the Euclidean norm over the generated positions (Section~\ref{sec:embeddings}).
The attractive drift \(V^{+}_j\) is the kernel-weighted barycenter of the
positives minus \(x_j\), which is \(y_1-x_j\) with a single positive; the
repulsive drift \(V^{-}_j\) is the same quantity for the other generated
samples of the same problem, which serve as its negatives as in the original
Drifting Model, and enters with a minus sign, so that it pushes \(x_j\) away
from them.
The weights of the two drifts, \((\lambda^{+},\lambda^{-})\), are set in Section~\ref{sec:experiments}, and Appendix~\ref{app:drift_field} gives the exact form of both terms.

\paragraph{Training through rounds without backpropagation through time.}
Each update samples the number of rounds and trains only the terminal round, running the earlier rounds with the gradient stopped (marked \(\operatorname{sg}\) in Figure~\ref{fig:reladrift_overview}), so that every round is trained without backpropagation through time.
Sampling the terminal round allows the shared network to be trained
at every depth across updates. Applying the loss at all rounds in
each update would require gradient computation at every round.
For each problem, we sample the number of rounds \(R\sim\operatorname{Uniform}\{1,\ldots,R_{\max}\}\), where the \emph{training depth} \(R_{\max}\) is the largest number of rounds used in training.
The model runs the rounds \(r=0,\ldots,R-1\), the last of which, round \(R-1\), is the \emph{terminal round}.
In the source formulation, each generated sample \(x_j\) comes from its own noise \(\epsilon_j\sim p_{\mathrm{init}}\) through the rounds
\[
\begin{aligned}
d_1&=\operatorname{sg}\bigl(f_\theta(\epsilon_j,P,0)\bigr),\\
d_{r+1}&=\operatorname{sg}\bigl(f_\theta(d_r,P,r)\bigr)
\quad (r=1,\ldots,R-2),\\
x_j&=f_\theta(d_{R-1},P,R-1),
\end{aligned}
\]
with \(x_j=f_\theta(\epsilon_j,P,0)\) when \(R=1\).
Only the terminal round's activations are stored, whatever \(R\) is, and the earlier rounds cost only forward passes (Appendix~\ref{app:training_cost}).
Because the terminal round is sampled, every refinement round \(r=1,\ldots,R_{\max}-1\) is trained on the drafts that the current model itself produces for it, as in training on model-generated inputs~\citep{bengio2015scheduled,savinov2022sundae}, and the first round is trained on noise, as the Drifting Model is.

The \(n\) generated samples of a problem branch at the first round in the source
formulation and at the terminal round in the additive and conditional
formulations, so that they are different drafts.
In the additive and conditional formulations, the recurrence is the same with
\(f_\theta(d_r,P,r)\) replaced by \(f_\theta(\epsilon_r,d_r,P,r)\), where a new
noise \(\epsilon_r\sim p_{\mathrm{init}}\) is drawn at every round \(r\) and the
trajectory starts from the initial draft \(d_0\); one such trajectory is shared
by the generated samples up to round \(R-2\), and \(n\) independent noises at
the terminal round give the \(n\) generated samples.
Appendix~\ref{app:training_procedure} describes one training step in detail (Algorithm~\ref{alg:iterative_training}).

\subsection{Generation and Test-Time Scaling}\label{sec:generation}

At test time, the number of rounds, the NFE \(N\), is chosen freely and is independent of the number of rounds \(R\) and the training depth \(R_{\max}\) used in training.
Generation with NFE \(=N\) runs the rounds \(r=0,\ldots,N-1\): the prompt is encoded once, the draft is updated \(N\) times with the same model \(f_\theta\), and the decoder is applied once to the final draft (Algorithm~\ref{alg:generation} in Appendix~\ref{app:generation_procedure}).

\paragraph{Rounds beyond the training depth.}
Generation can run more rounds than the training depth: every refinement round
(\(r\geq1\)) is trained to correct the draft that it receives, so repeating
this correction beyond \(R_{\max}\) is expected to keep improving the draft;
Section~\ref{sec:experiments} tests this expectation.
The round embedding \(\phi(r)\), however, is seen in training only for
\(r<R_{\max}\), so we clamp it to the
last trained round,
\[
\phi(r)\;\to\;\phi\bigl(\min(r,\,R_{\max}-1)\bigr);
\]
since all refinement rounds share every parameter of \(f_\theta\), each
additional round then applies exactly the map trained at round \(R_{\max}-1\),
and the number of rounds becomes a pure test-time choice.
The clamp is part of the model definition; Appendix~\ref{app:clamp} reports the
results without it.

\section{Experiments}\label{sec:experiments}

We evaluate how ReLaDrift scales with inference-time computation
on Sudoku Hard (Section~\ref{sec:sudoku}) and GSM8K (Section~\ref{sec:gsm8k}),
comparing it with an iterated one-step Drifting Model and established
diffusion and flow language models.
We train all compared methods on the same task-specific data,
largely following the setups of S-FLM~\citep{deschenaux2026sflm}.
Differences from the published setups and comparisons with reported
baseline results are detailed in Appendix~\ref{app:settings}.

\paragraph{Evaluation protocol.}\label{sec:protocol}
We measure accuracy across a range of inference budgets,
using the number of function evaluations (NFE) as a common measure
while keeping model weights fixed.
For ReLaDrift, each round counts as one NFE, with the prompt encoded once and only the final draft decoded.
For other parallel generators, NFE is the number of ODE or sampling steps (S-FLM, Duo, MDLM, and FRM) or of refinement rounds (FMLM+);
the one-step Drifting Model is iterated by decoding and re-encoding
its outputs between rounds (Appendix~\ref{app:naive_recurrence}).
We evaluate a single generation trajectory per problem using EMA weights,
without inference-time verification or candidate selection;
GSM8K programs are executed only to compute accuracy.
Since NFE does not capture differences in per-step cost,
we also compare wall-clock generation time on GSM8K.
\paragraph{Training configuration.}
Unless otherwise stated, we use $n=2$ generated samples per problem
to reduce the cost of recurrent training and, on GSM8K, verification.
To stabilize training with few generated samples,
we weight the attractive and repulsive drifts as
$\lambda^{+}V^{+}-\lambda^{-}V^{-}$ with
$(\lambda^{+},\lambda^{-})=(2,1)$,
strengthening the attraction toward target samples.
We analyze these choices in Appendix~\ref{app:lambda}
and their computational costs in Appendix~\ref{app:training_cost}.

\begin{table*}[t]
\centering
\small
\setlength{\tabcolsep}{4pt}
\resizebox{\textwidth}{!}{%
\begin{tabular}{lrrrrrrrrr}
\toprule
Method & NFE=1 & 4 & 5 & 8 & 16 & 32 & 64 & 128 & 180 \\
\midrule
\fs AR (sequential) &
\multicolumn{9}{c}{\fs Greedy: \(0.5\pm0.1\); sampling: \(0.5\pm0.2\)} \\
\midrule
\fs MDLM
& \fs \(0.6\pm0.2\) & \fs \(8.1\pm1.0\) & \fs \(11.6\pm1.6\)
& \fs \(15.9\pm1.9\) & \fs \(22.7\pm2.6\) & \fs \(27.1\pm2.1\)
& \fs \(30.1\pm2.2\) & \fs \(32.2\pm1.6\) & \fs \(32.6\pm2.6\) \\
\fs Duo
& \fs \(4.0\pm2.1\) & \fs \(55.2\pm6.7\) & \fs \(56.8\pm5.6\)
& \fs \(60.6\pm5.4\) & \fs \(61.9\pm4.9\) & \fs \(62.9\pm4.7\)
& \fs \(61.8\pm4.8\) & \fs \(62.9\pm4.8\) & \fs \(61.5\pm4.9\) \\
\fs S-FLM
& \fs \(6.7\pm1.1\) & \fs \(33.0\pm2.9\) & \fs \(34.9\pm3.2\)
& \fs \(38.0\pm3.0\) & \fs \(40.6\pm3.2\) & \fs \(41.9\pm3.2\)
& \fs \(42.5\pm3.4\) & \fs \(43.0\pm3.3\) & \fs \(43.2\pm3.3\) \\
\fs FMLM+
& \fs \(2.4\pm0.4\) & \fs \(67.0\pm2.4\) & \fs \(71.2\pm1.5\)
& \fs \(76.8\pm1.4\) & \fs \(77.5\pm2.1\) & \fs \(78.5\pm1.4\)
& \fs \(78.1\pm2.0\) & \fs \(78.5\pm1.5\) & \fs \(77.2\pm1.4\) \\
\fs FRM (Euler)
& \fs \(6.4\pm0.9\) & \fs \(54.9\pm1.0\) & \fs \(60.6\pm0.8\)
& \fs \(66.7\pm1.1\) & \fs \(71.8\pm1.5\) & \fs \(74.2\pm1.6\)
& \fs \(75.2\pm1.7\) & \fs \(75.8\pm1.7\) & \fs \(75.9\pm1.6\) \\
\fs FRM (SDE)
& \fs \(6.4\pm0.9\) & \fs \(55.5\pm0.6\) & \fs \(61.2\pm1.1\)
& \fs \(67.3\pm0.9\) & \fs \(72.6\pm1.3\) & \fs \(75.1\pm1.6\)
& \fs \(77.0\pm2.1\) & \fs \(78.0\pm2.7\) & \fs \(79.4\pm2.0\) \\
\fs FRM (held-time)
& \fs \(6.4\pm0.9\) & \fs \(61.0\pm2.5\) & \fs \(65.3\pm1.4\)
& \fs \(70.6\pm1.5\) & \fs \(73.5\pm1.7\) & \fs \(74.4\pm1.5\)
& \fs \(74.5\pm1.5\) & \fs \(74.5\pm1.5\) & \fs \(74.5\pm1.5\) \\
\fs DFM
& \fs \(6.3\pm4.5\) & \fs \(25.8\pm7.2\) & \fs \(26.2\pm7.0\)
& \fs \(27.9\pm6.8\) & \fs \(29.2\pm6.5\) & \fs \(29.6\pm6.5\)
& \fs \(30.1\pm6.3\) & \fs \(30.3\pm6.4\) & \fs \(30.4\pm6.4\) \\
\midrule
Drifting Model
& \(5.1\pm0.8\) & \(4.9\pm0.8\) & \(5.0\pm0.8\)
& \(4.9\pm0.8\) & \(4.9\pm0.8\) & \(4.9\pm0.8\)
& \(4.9\pm0.8\) & \(4.9\pm0.8\) & \(4.9\pm0.8\) \\
\textbf{ReLaDrift (ours)}
& \textbf{7.3\(\pm\)1.0}
& \textbf{78.9\(\pm\)0.6}
& \textbf{82.5\(\pm\)0.8}
& \textbf{88.4\(\pm\)0.4}
& \textbf{92.1\(\pm\)0.4}
& \textbf{93.1\(\pm\)0.6}
& \textbf{93.5\(\pm\)0.7}
& \textbf{93.9\(\pm\)0.7}
& \textbf{94.0\(\pm\)0.7} \\
\bottomrule
\end{tabular}}
\caption{
\textbf{Sudoku Hard accuracy (\%) across inference budgets.}
Exact-match accuracy on 2k validation puzzles after 20k
training updates (mean \(\pm\) standard deviation over three seeds).
ReLaDrift uses the Source formulation with \(R_{\max}=5\).
The Drifting Model is iterated by decoding and re-encoding
its output between rounds (Appendix~\ref{app:naive_recurrence}).
FRM and DFM are our implementations (Appendix~\ref{app:settings}).
Bold indicates the best accuracy at each NFE.
}
\label{tab:sudoku_results}
\end{table*}

\begin{table*}[t]
\centering
\small
\setlength{\tabcolsep}{4pt}
\resizebox{\textwidth}{!}{%
\begin{tabular}{lrrrrrrrrr}
\toprule
Setting & NFE=1 & 4 & 5 & 8 & 16 & 32 & 64 & 128 & 180 \\
\midrule
\textbf{Default}
& \(7.3\pm1.0\) & \(78.9\pm0.6\) & \(82.5\pm0.8\)
& \(88.4\pm0.4\) & \(92.1\pm0.4\) & \(93.1\pm0.6\)
& \(93.5\pm0.7\) & \(93.9\pm0.7\) & \(94.0\pm0.7\) \\
\midrule
\multicolumn{10}{l}{\textbf{\textit{Input formulation} (Default: Source)}} \\
\quad- Additive
& \(7.1\pm2.2\) & \(77.6\pm2.6\) & \(81.5\pm3.1\)
& \(86.2\pm3.6\) & \(89.5\pm3.1\) & \(90.5\pm3.0\)
& \(91.0\pm2.8\) & \(91.6\pm2.5\) & \(91.6\pm2.8\) \\
\quad- Conditional
& \(7.4\pm0.9\) & \(77.6\pm1.7\) & \(81.1\pm1.1\)
& \(86.1\pm1.2\) & \(89.6\pm1.8\) & \(90.7\pm1.7\)
& \(91.2\pm2.0\) & \(91.6\pm2.3\) & \(91.7\pm2.2\) \\
\midrule
\multicolumn{10}{l}{\textbf{\textit{Training depth \(R_{\max}\)} (Default: $5$)}} \\
\quad- \(3\)
& \(5.7\pm2.7\) & \(66.2\pm3.4\) & \(68.5\pm2.9\)
& \(71.4\pm2.1\) & \(72.6\pm2.0\) & \(72.9\pm1.9\)
& \(73.1\pm1.9\) & \(73.1\pm2.0\) & \(73.2\pm2.0\) \\
\quad- \(8\)
& \(6.4\pm2.0\) & \(83.4\pm1.7\) & \(86.9\pm1.7\)
& \(91.2\pm1.7\) & \(94.9\pm1.3\) & \(96.0\pm1.0\)
& \(96.6\pm1.0\) & \(96.8\pm0.9\) & \(96.9\pm0.8\) \\
\quad- \(16\)
& \(4.2\pm2.1\) & \(85.2\pm2.9\) & \(88.8\pm2.1\)
& \(93.0\pm1.7\) & \(96.1\pm0.8\) & \(97.2\pm0.5\)
& \(97.4\pm0.4\) & \(97.5\pm0.4\) & \(97.6\pm0.4\) \\
\bottomrule
\end{tabular}
}
\caption{
\textbf{Ablation of ReLaDrift on Sudoku Hard.}
Exact-match accuracy (\%; mean \(\pm\) standard deviation over three
seeds) under the settings of Table~\ref{tab:sudoku_results}.
The default Source formulation uses \(R_{\max}=5\) and serves
as the reference for both the input-formulation and
training-depth comparisons.
}
\label{tab:sudoku_ablations}
\end{table*}

\subsection{Sudoku}\label{sec:sudoku}
\paragraph{Experimental setup.}
We evaluate ReLaDrift on Sudoku Hard from the S-FLM
benchmark~\citep{deschenaux2026sflm}, comprising 48,000 training
and 2,000 validation puzzles generated using
\citet{alp2024sudoku}, each with a unique solution and 30 of
81 digits visible.
Models generate complete solutions, evaluated by exact match
(Appendix~\ref{app:encoding}).
We compare against AR, MDLM~\citep{sahoo2024mdlm},
Duo~\citep{sahoo2025duo}, S-FLM,
FMLM+~\citep{agarwal2026posterior},
FRM~\citep{helbling2026frm}, and
DFM~\citep{ma2026dfm}, all trained on the same data for
20,000 updates following the S-FLM optimization settings.
We reimplement FRM and DFM for this benchmark.
All models have comparable sizes (25.2--28.6M parameters;
28.4M for ReLaDrift).
ReLaDrift uses one-hot representations with $\arg\max$ decoding,
without a pretrained encoder or verifier.
We also train a one-step Drifting Model with matched architecture,
loss, data, and training budget to see whether naive repetition
alone yields test-time scaling.
Further details are in Appendix~\ref{app:settings}.

\paragraph{Main Results}
ReLaDrift exhibits strong test-time scaling on Sudoku Hard,
whereas simply iterating the one-step Drifting Model yields no improvement
(Table~\ref{tab:sudoku_results} and Figure~\ref{fig:teaser}, top).
Despite being trained with at most $R_{\max}=5$ rounds,
its accuracy improves monotonically with NFE across all three seeds,
reaching 94.0\% at NFE=180.
ReLaDrift outperforms all compared diffusion and flow language models
across NFEs.
These results establish recurrent drifting as an effective means of trading inference-time computation for accuracy.

\paragraph{Ablations.}
Table~\ref{tab:sudoku_ablations} examines the input formulation (\emph{Source, Additive, Conditional}) and training depth $R_{\max}$.
All three formulations exhibit strong test-time scaling, suggesting that recurrent drifting is robust to how previous drafts are incorporated.
Among them, Source (default) achieves the highest accuracy, with the lowest seed variability at NFE $\geq 8$. 
Increasing the training depth further improves multi-round accuracy (NFE $\geq 4$), even beyond the training depth $R_{\max}=5$.
This improvement comes at the cost of additional forward passes during training, while backpropagation remains limited to
the terminal round (Sec.~\ref{sec:method}, Appendix~\ref{app:training_cost}).

\subsection{GSM8K}\label{sec:gsm8k}

\begin{table*}[t]
\centering
\small
\begin{tabular}{lrrrrrrrrrrr}
\toprule
Method & \multicolumn{11}{c}{NFE} \\
\cmidrule(lr){2-12}
 & 1 & 4 & 5 & 8 & 16 & 32 & 64 & 128 & 256 & 512 & 1024 \\
\midrule
\fs AR (sequential) &
\multicolumn{11}{c}{\fs 72.3} \\
\midrule
\fs MDLM
& \fs 0.3 & \fs 1.3 & \fs 1.9 & \fs 5.2 & \fs 12.1 & \fs 18.8
& \fs 27.2 & \fs 28.7 & \fs 29.7 & \fs 32.0 & \fs 32.1 \\
\fs Duo
& \fs 0.5 & \fs 5.2 & \fs 6.8 & \fs 12.5 & \fs 23.1 & \fs \textbf{30.6}
& \fs \textbf{34.7} & \fs \textbf{34.9} & \fs \textbf{36.3}
& \fs \textbf{37.1} & \fs \textbf{35.9} \\
\fs S-FLM
& \fs 0.5 & \fs 10.5 & \fs 13.3 & \fs 16.2 & \fs 17.7 & \fs 19.9
& \fs 20.2 & \fs 20.5 & \fs 20.3 & \fs 20.4 & \fs 20.8 \\
\fs FMLM+
& \fs 0.0 & \fs 2.6 & \fs 4.9 & \fs 10.1 & \fs 20.5 & \fs 24.9
& \fs 26.0 & \fs 25.6 & \fs 25.5 & \fs 25.5 & \fs 25.5 \\
\midrule
Drifting Model
& 8.4 & 7.8 & 7.7 & 7.7 & 7.7 & 7.7
& 7.7 & 7.7 & 7.7 & 7.7 & 7.7 \\
\textbf{ReLaDrift}
& \textbf{13.2} & \textbf{27.3} & \textbf{27.4}
& \textbf{29.0} & \textbf{29.4} & 29.4
& 29.4 & 29.5 & 29.5 & 29.4 & 29.4 \\
\bottomrule
\end{tabular}%
\caption{
\textbf{GSM8K test accuracy (\%) across inference budgets.}
Accuracy on 1,319 test problems after 250,000 training updates
(single training seed; one generated program per problem).
ReLaDrift uses $K=12$ given positives and $R_{\max}=5$.
The Drifting Model is iterated by decoding and re-encoding
its output between rounds (Appendix~\ref{app:naive_recurrence}).
AR, MDLM, Duo, and S-FLM use $T=0.1$; FMLM+ uses $\arg\max$.
Bold indicates the best parallel generator at each NFE.
}
\label{tab:gsm8k_results}
\end{table*}

\paragraph{General setup.}
Following S-FLM~\citep{deschenaux2026sflm}, we train on
TinyGSM~\citep{liu2023tinygsm}, a dataset of synthetic math problems
paired with Python solutions, and evaluate on the 1,319 GSM8K
test problems~\citep{cobbe2021gsm8k}.
Accuracy is measured by executing one generated program per problem
and comparing its output with the ground-truth answer.
To provide multiple reference solutions per problem, we use
1,588,781 TinyGSM problems, each paired with 12 programs generated
by Qwen3.5-9B~\citep{qwen2026qwen35}, without filtering for correctness
(Appendix~\ref{app:qwen_prompt}).
All methods use the same problems; the compared methods train on
all 12 programs, while ReLaDrift uses the first $K\in\{1,4,12\}$
as its given positives.

\paragraph{Models and training.}
We compare with AR, MDLM, Duo, S-FLM, FMLM+, and a separately trained one-step Drifting Model.
All methods are trained for 250,000 updates with batch size 512, following the S-FLM optimization settings.
The diffusion and flow baselines use about 160M parameters, while ReLaDrift uses 164.9M in total, including a frozen CodeT5-small encoder~\citep{wang2021codet5} and a separately trained, frozen decoder.
ReLaDrift operates in the encoder's contextualized latent space, with the decoder mapping the final draft to tokens in parallel.
The one-step Drifting Model uses the same representation, drifting objective, verifier, and training settings as ReLaDrift, but without recurrent refinement.
AR, S-FLM, Duo, and MDLM use sampling temperature $T=0.1$, and FMLM+ uses $\arg\max$ decoding.
Implementation details are in Appendix~\ref{app:settings}.

\paragraph{Training with multiple valid solutions.}\label{sec:verifier}
Unlike Sudoku, GSM8K admits multiple correct programs for each problem.
Because the drifting objective attracts generated samples toward
the given positives, valid programs not represented by these
references may be pulled toward different solutions.
To address this, we augment the $K$ given positives with
model-generated programs that pass execution-based verification.
The verifier compares their outputs against a pseudo-reference
obtained by voting over the answers of the $K$ given programs,
rather than using the ground-truth answer.
Verified programs enter a per-problem positive queue with up to
four additional entries and are excluded from the repulsion
applied to other generated samples.
Verification and the queue are used only during training
(Appendix~\ref{app:verified_step}).

\begin{figure}[t]
\centering
\includegraphics[width=\columnwidth]{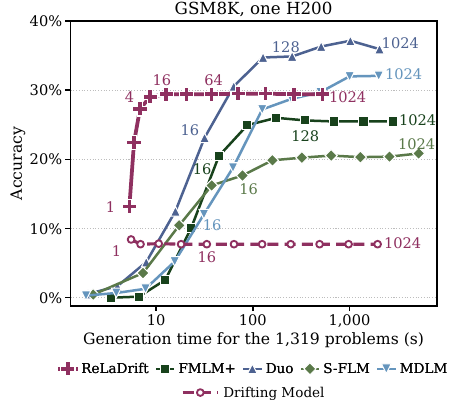}
\caption{\textbf{GSM8K test accuracy versus the wall-clock time of generating the 1,319 test programs on one H200} (log scale; batches of 64 problems; settings of Table~\ref{tab:gsm8k_results}). The labels are the NFE; the time of ReLaDrift includes the encoder and the decoder.}
\label{fig:walltime}
\end{figure}

\paragraph{Main results.}
ReLaDrift also achieves test-time scaling on GSM8K, demonstrating
that recurrent drifting benefits from additional inference rounds
even with multiple valid solutions
(Table~\ref{tab:gsm8k_results} and Figure~\ref{fig:teaser}, bottom).
Unlike the one-step Drifting Model, which does not improve when iterated,
ReLaDrift's accuracy increases from 13.2\% at NFE=1 to 29.4\% at NFE=16,
even beyond its training depth $R_{\max}=5$.
Compared with established diffusion and flow language models,
ReLaDrift achieves the highest accuracy at matched NFE up to 16
and remains competitive at larger inference budgets, where its
accuracy eventually saturates.
Together with the Sudoku results, these findings position recurrent
drifting as a competitive approach to test-time-scalable parallel
language generation.

\paragraph{Inference efficiency.}
ReLaDrift is also more efficient per function evaluation,
taking 24 ms per round versus 95 ms per step for Duo on a single H200 GPU
(Appendix~\ref{app:training_cost}).
Consequently, ReLaDrift achieves 29.4\% accuracy in just 13 seconds,
whereas Duo requires 63 seconds to reach a comparable accuracy of 30.6\%
(Figure~\ref{fig:walltime}).
This advantage is more pronounced in wall-clock time than in
matched-NFE comparisons.

\begin{table}[t]
\centering
\footnotesize
\setlength{\tabcolsep}{4pt}
\begin{tabular}{lrrrr}
\toprule
Setting & \multicolumn{4}{c}{NFE} \\
\cmidrule(lr){2-5}
 & 1 & 4 & 8 & 16 \\
\midrule
\textbf{Default}
& 13.2 & 27.3 & 29.0 & 29.4 \\
\midrule
\multicolumn{5}{l}{\textit{\textbf{Number of given positives}} $K$ (default 12)} \\
\quad- $4$  & 10.4 & 22.7 & 24.4 & 24.9 \\
\quad- $1$  & 1.0  & 8.7  & 10.7 & 12.2 \\
\midrule
\multicolumn{5}{l}{\textit{\textbf{Training depth}} $R_{\max}$ (default 5)} \\
\quad- $8$  & 10.2 & 24.2 & 26.2 & 27.7 \\
\quad- $16$ & 8.0  & 24.0 & 26.8 & 27.2 \\
\midrule
\multicolumn{5}{l}{\textit{\textbf{Training components}}} \\
\quad- Without verifier and queue
& 4.5 & 17.0 & 18.5 & 20.0 \\
\midrule
\multicolumn{5}{l}{\textit{\textbf{Encoder}} (Default: contextualized CodeT5.)} \\
\quad- Uncontextualized CodeT5.
& 6.7 & 15.1 & 16.7 & 17.7 \\
\quad- Random emb. table
& 1.4 & 5.5 & 6.7 & 7.1 \\
\bottomrule
\end{tabular}
\caption{
\textbf{Ablation of ReLaDrift on GSM8K.}
Test accuracy (\%) after 250,000 training updates.
The default uses $K=12$, $R_{\max}=5$, a contextualized
CodeT5 encoder, and training-time verification with a positive queue.
Embedding-table variants replace the contextualized encoder
with context-free representations.
Additional settings and results across all NFE are provided
in Appendix~\ref{app:gsm8k_ablation}.
}
\label{tab:gsm8k_main_ablation}
\end{table}

\paragraph{Ablations.}
Unlike Sudoku, GSM8K involves multiple valid solutions and requires additional design choices for training in a contextualized latent space.
We therefore examine three GSM8K-specific factors: the number of given positives $K$, training-time verification, and the encoder representation (Table~\ref{tab:gsm8k_main_ablation}).
Increasing $K$ substantially improves accuracy, while the verifier and positive queue and contextualized CodeT5 embeddings provide further gains.
Importantly, test-time scaling persists even without these components, indicating that they improve accuracy rather than enable recurrent refinement.
Notably, unlike Sudoku, where increasing the training depth $R_{\max}$ improves multi-round accuracy, GSM8K performs best at $R_{\max}=5$, although both tasks benefit from inference rounds beyond their training depths.
Increasing $R_{\max}$ exposes the model to longer refinement trajectories but reduces the frequency of supervision at each depth under a fixed training budget, suggesting that the balance between these effects is task-dependent.
Further ablations of the input formulation, drift-field coefficients, and round-embedding clamp are in Appendices~\ref{app:gsm8k_ablation} and~\ref{app:clamp}.

\section{Conclusion}\label{sec:conclusion}

We introduced ReLaDrift, which turns the one-step Drifting Model into a recurrent language generator by iteratively refining its own drafts in a continuous latent space.
The model learns refinement by applying the drifting loss at a sampled terminal round, avoiding backpropagation through time.
Unlike the original Drifting Model, which does not improve when repeatedly applied, ReLaDrift exhibits test-time scaling on Sudoku Hard and GSM8K, enabling the computation--quality trade-off of established diffusion and flow language models.
It outperforms these models across NFE=4--180 on Sudoku Hard and up to NFE=16 on GSM8K, positioning drifting-based generation as a promising alternative for test-time-scalable parallel language generation.

\paragraph{Discussion}
ReLaDrift incurs approximately $3.3\times$ the per-update
training cost of a one-pass method
(Appendix~\ref{app:training_cost}).
This overhead reflects both recurrent forward passes and a
broader challenge inherent to Drifting Models: estimating
the drift field requires multiple generated samples,
each involving additional model evaluations.
We limit this cost by using only $n=2$ generated samples
per problem.
Improving the computational efficiency of drift-field estimation
is an important research direction in its own right,
beyond the scope of this work, which focuses on enabling
test-time scaling for drifting-based generation.

\section*{Limitations}

On GSM8K, our advantage over parallel generators depends on a verifier and pretrained CodeT5 embeddings, while test-time scaling persists without these components (Sec.~\ref{sec:gsm8k}).
The verifier requires executable outputs and relies on
voted pseudo-references that may be incorrect.
Furthermore, GSM8K results use a single training seed due to limited training resources.
Evaluating broader tasks and model scales, improving
verifier-free training with multiple valid targets, and
narrowing the gap to autoregressive models remain future work.

\section*{Acknowledgments}

This research was conducted by AIST and Institute of Science Tokyo under FRONTia, a Japanese national program led by the Ministry of Economy, Trade and Industry of Japan (METI) and the New Energy and Industrial Technology Development Organization (NEDO), aimed at developing a domestic multimodal foundation model for AI robots and physical AI.
\bibliography{custom}

@article{agarwal2026posterior,
    title={Posterior Refinement: Fast Language Generation via Any-Order Flow Maps},
    author={Manan Agarwal and Sheel Shah and Chanhyuk Lee
            and Jaehoon Yoo and Jerry Huang and Seunghoon Hong
            and Aditi Raghunathan and Jinwoo Kim and Nicholas M. Boffi},
    journal={arXiv preprint arXiv:2606.24773},
    year={2026},
    url = {https://arxiv.org/abs/2606.24773}
}

@misc{allal2024smollm,
  title = {{SmolLM} - blazingly fast and remarkably powerful},
  author = {Ben Allal, Loubna and Lozhkov, Anton and Bakouch, Elie},
  year = {2024},
  month = {jul},
  howpublished = {Hugging Face blog, \url{https://huggingface.co/blog/smollm}},
  note = {Model card: \url{https://huggingface.co/HuggingFaceTB/SmolLM-135M}}
}

@misc{alp2024sudoku,
  title = {Sudoku Puzzle Generator},
  author = {Alp, Ali},
  howpublished = {\url{https://github.com/alicommit-malp/sudoku}},
  year = {2024},
  note = {GitHub repository, created September 2024}
}

@inproceedings{austin2021d3pm,
 author = {Austin, Jacob and Johnson, Daniel D. and Ho, Jonathan and Tarlow, Daniel and van den Berg, Rianne},
 booktitle = {Advances in Neural Information Processing Systems},
 editor = {M. Ranzato and A. Beygelzimer and Y. Dauphin and P.S. Liang and J. Wortman Vaughan},
 pages = {17981--17993},
 publisher = {Curran Associates, Inc.},
 title = {Structured Denoising Diffusion Models in Discrete State-Spaces},
 url = {https://proceedings.neurips.cc/paper_files/paper/2021/file/958c530554f78bcd8e97125b70e6973d-Paper.pdf},
 volume = {34},
 year = {2021}
}

@inproceedings{bai2019deq,
 author = {Bai, Shaojie and Kolter, J. Zico and Koltun, Vladlen},
 booktitle = {Advances in Neural Information Processing Systems},
 editor = {H. Wallach and H. Larochelle and A. Beygelzimer and F. d\textquotesingle Alch\'{e}-Buc and E. Fox and R. Garnett},
 pages = {},
 publisher = {Curran Associates, Inc.},
 title = {Deep Equilibrium Models},
 url = {https://proceedings.neurips.cc/paper_files/paper/2019/file/01386bd6d8e091c2ab4c7c7de644d37b-Paper.pdf},
 volume = {32},
 year = {2019}
}

@inproceedings{bengio2015scheduled,
 author = {Bengio, Samy and Vinyals, Oriol and Jaitly, Navdeep and Shazeer, Noam},
 booktitle = {Advances in Neural Information Processing Systems},
 editor = {C. Cortes and N. Lawrence and D. Lee and M. Sugiyama and R. Garnett},
 pages = {},
 publisher = {Curran Associates, Inc.},
 title = {Scheduled Sampling for Sequence Prediction with Recurrent Neural Networks},
 url = {https://proceedings.neurips.cc/paper_files/paper/2015/file/e995f98d56967d946471af29d7bf99f1-Paper.pdf},
 volume = {28},
 year = {2015}
}

@article{boffi2025flowmap,
  title={Flow map matching with stochastic interpolants: A mathematical framework for consistency models},
  author={Nicholas Matthew Boffi and Michael Samuel Albergo and Eric Vanden-Eijnden},
  journal={Transactions on Machine Learning Research},
  issn={2835-8856},
  year={2025},
  url={https://openreview.net/forum?id=cqDH0e6ak2},
  note={}
}

@misc{cobbe2021gsm8k,
      title={Training Verifiers to Solve Math Word Problems},
      author={Karl Cobbe and Vineet Kosaraju and Mohammad Bavarian and Mark Chen and Heewoo Jun and Lukasz Kaiser and Matthias Plappert and Jerry Tworek and Jacob Hilton and Reiichiro Nakano and Christopher Hesse and John Schulman},
      year={2021},
      eprint={2110.14168},
      archivePrefix={arXiv},
      primaryClass={cs.LG},
      url={https://arxiv.org/abs/2110.14168},
}

@inproceedings{
dehghani2019universal,
title={Universal Transformers},
author={Mostafa Dehghani and Stephan Gouws and Oriol Vinyals and Jakob Uszkoreit and Lukasz Kaiser},
booktitle={International Conference on Learning Representations},
year={2019},
url={https://openreview.net/forum?id=HyzdRiR9Y7},
}

@inproceedings{deng2026generativemodelingdrifting,
  title={Generative Modeling via Drifting},
  author={Deng, Mingyang and Li, He and Li, Tianhong and Du, Yilun and He, Kaiming},
  booktitle={Advances in Neural Information Processing Systems (NeurIPS)},
  year={2026}
}

@inproceedings{deschenaux2025sdtt,
 author = {Deschenaux, Justin and Gulcehre, Caglar},
 booktitle = {International Conference on Learning Representations},
 editor = {Y. Yue and A. Garg and N. Peng and F. Sha and R. Yu},
 pages = {5007--5045},
 title = {{Beyond Autoregression: Fast LLMs via Self-Distillation Through Time}},
 url = {https://proceedings.iclr.cc/paper_files/paper/2025/file/0f93c3e9b557980d93016671acd94bd2-Paper-Conference.pdf},
 volume = {2025},
 year = {2025}
}

@misc{deschenaux2026sflm,
      title={Language Modeling with Hyperspherical Flows},
      author={Justin Deschenaux and Caglar Gulcehre},
      year={2026},
      eprint={2605.11125},
      archivePrefix={arXiv},
      primaryClass={cs.LG},
      url={https://arxiv.org/abs/2605.11125},
}

@inproceedings{frans2025shortcut,
 author = {Frans, Kevin and Hafner, Danijar and Levine, Sergey and Abbeel, Pieter},
 booktitle = {International Conference on Learning Representations},
 editor = {Y. Yue and A. Garg and N. Peng and F. Sha and R. Yu},
 pages = {34668--34684},
 title = {One Step Diffusion via Shortcut Models},
 url = {https://proceedings.iclr.cc/paper_files/paper/2025/file/559a0998fab1d19b80e7e43a5852401c-Paper-Conference.pdf},
 volume = {2025},
 year = {2025}
}

@InProceedings{gao2023pal,
  title = 	 {{PAL}: Program-aided Language Models},
  author =       {Gao, Luyu and Madaan, Aman and Zhou, Shuyan and Alon, Uri and Liu, Pengfei and Yang, Yiming and Callan, Jamie and Neubig, Graham},
  booktitle = 	 {Proceedings of the 40th International Conference on Machine Learning},
  pages = 	 {10764--10799},
  year = 	 {2023},
  editor = 	 {Krause, Andreas and Brunskill, Emma and Cho, Kyunghyun and Engelhardt, Barbara and Sabato, Sivan and Scarlett, Jonathan},
  volume = 	 {202},
  series = 	 {Proceedings of Machine Learning Research},
  month = 	 {23--29 Jul},
  publisher =    {PMLR},
  url = 	 {https://proceedings.mlr.press/v202/gao23f.html},
}

@inproceedings{
geiping2025recurrent,
title={Scaling up Test-Time Compute with Latent Reasoning: A Recurrent Depth Approach},
author={Jonas Geiping and Sean Michael McLeish and Neel Jain and John Kirchenbauer and Siddharth Singh and Brian R. Bartoldson and Bhavya Kailkhura and Abhinav Bhatele and Tom Goldstein},
booktitle={The Thirty-ninth Annual Conference on Neural Information Processing Systems},
year={2025},
url={https://openreview.net/forum?id=S3GhJooWIC}
}

@inproceedings{
geng2025meanflow,
title={Mean Flows for One-step Generative Modeling},
author={Zhengyang Geng and Mingyang Deng and Xingjian Bai and J Zico Kolter and Kaiming He},
booktitle={The Thirty-ninth Annual Conference on Neural Information Processing Systems},
year={2025},
url={https://openreview.net/forum?id=uWj4s7rMnR}
}

@inproceedings{ghazvininejad2019maskpredict,
    title = "Mask-Predict: Parallel Decoding of Conditional Masked Language Models",
    author = "Ghazvininejad, Marjan  and
      Levy, Omer  and
      Liu, Yinhan  and
      Zettlemoyer, Luke",
    editor = "Inui, Kentaro  and
      Jiang, Jing  and
      Ng, Vincent  and
      Wan, Xiaojun",
    booktitle = "Proceedings of the 2019 Conference on Empirical Methods in Natural Language Processing and the 9th International Joint Conference on Natural Language Processing ({EMNLP}-{IJCNLP})",
    month = nov,
    year = "2019",
    address = "Hong Kong, China",
    publisher = "Association for Computational Linguistics",
    url = "https://aclanthology.org/D19-1633/",
    doi = "10.18653/v1/D19-1633",
    pages = "6112--6121",
}

@inproceedings{goodfellow2014gan,
 author = {Goodfellow, Ian J. and Pouget-Abadie, Jean and Mirza, Mehdi and Xu, Bing and Warde-Farley, David and Ozair, Sherjil and Courville, Aaron and Bengio, Yoshua},
 booktitle = {Advances in Neural Information Processing Systems},
 editor = {Z. Ghahramani and M. Welling and C. Cortes and N. Lawrence and K. Weinberger},
 pages = {},
 publisher = {Curran Associates, Inc.},
 title = {Generative Adversarial Nets},
 url = {https://proceedings.neurips.cc/paper_files/paper/2014/file/f033ed80deb0234979a61f95710dbe25-Paper.pdf},
 volume = {27},
 year = {2014}
}

@inproceedings{gu2019levenshtein,
 author = {Gu, Jiatao and Wang, Changhan and Zhao, Junbo},
 booktitle = {Advances in Neural Information Processing Systems},
 editor = {H. Wallach and H. Larochelle and A. Beygelzimer and F. d\textquotesingle Alch\'{e}-Buc and E. Fox and R. Garnett},
 pages = {},
 publisher = {Curran Associates, Inc.},
 title = {Levenshtein Transformer},
 url = {https://proceedings.neurips.cc/paper_files/paper/2019/file/675f9820626f5bc0afb47b57890b466e-Paper.pdf},
 volume = {32},
 year = {2019}
}

@misc{helbling2026frm,
      title={Flow Reasoning Models: Turning Flows Into Efficient Recurrent Reasoners},
      author={Alec Helbling and Andrey Bryutkin and Mauro Martino and Duen Horng Chau and Nima Dehmamy and Hendrik Strobelt},
      year={2026},
      eprint={2606.29150},
      archivePrefix={arXiv},
      primaryClass={cs.AI},
      url={https://arxiv.org/abs/2606.29150},
}

@misc{hu2026elf,
      title={{ELF: Embedded Language Flows}},
      author={Keya Hu and Linlu Qiu and Yiyang Lu and Hanhong Zhao and Tianhong Li and Yoon Kim and Jacob Andreas and Kaiming He},
      year={2026},
      eprint={2605.10938},
      archivePrefix={arXiv},
      primaryClass={cs.CL},
      url={https://arxiv.org/abs/2605.10938},
}

@inproceedings{lee2018iterative,
  title = {Deterministic Non-Autoregressive Neural Sequence Modeling by Iterative Refinement},
  author = {Lee, Jason  and Mansimov, Elman  and Cho, Kyunghyun},
  booktitle = {Proceedings of the 2018 Conference on Empirical Methods in Natural Language Processing},
  month = oct # "-" # nov,
  year = {2018},
  address = {Brussels, Belgium},
  publisher = {Association for Computational Linguistics},
  url = {https://aclanthology.org/D18-1149/},
  doi = {10.18653/v1/D18-1149},
  pages = {1173--1182}
}

@article{lee2026fmlm,
  title = {Flow Map Language Models: One-step Language Modeling via Continuous Denoising},
  author = {Lee, Chanhyuk and Yoo, Jaehoon and Agarwal, Manan and Shah, Sheel and Huang, Jerry and Raghunathan, Aditi and Hong, Seunghoon and Boffi, Nicholas M. and Kim, Jinwoo},
  journal = {arXiv preprint arXiv:2602.16813},
  year = {2026},
  url = {https://arxiv.org/abs/2602.16813}
}

@inproceedings{li2022diffusionlm,
  title = {{Diffusion-LM Improves Controllable Text Generation}},
  author = {Li, Xiang Lisa and Thickstun, John and Gulrajani, Ishaan and Liang, Percy and Hashimoto, Tatsunori B.},
  booktitle = {Advances in Neural Information Processing Systems},
  volume = {35},
  year = {2022},
  url = {https://proceedings.neurips.cc/paper_files/paper/2022/hash/1be5bc25d50895ee656b8c2d9eb89d6a-Abstract-Conference.html}
}

@article{liu2023tinygsm,
  title = {{TinyGSM}: achieving {$>$80\%} on {GSM8k} with small language models},
  author = {Liu, Bingbin and Bubeck, Sebastien and Eldan, Ronen and Kulkarni, Janardhan and Li, Yuanzhi and Nguyen, Anh and Ward, Rachel and Zhang, Yi},
  journal = {arXiv preprint arXiv:2312.09241},
  year = {2023},
  url = {https://arxiv.org/abs/2312.09241}
}

@inproceedings{lovelace2023ld4lg,
  title = {{Latent Diffusion for Language Generation}},
  author = {Lovelace, Justin and Kishore, Varsha and Wan, Chao and Shekhtman, Eliot and Weinberger, Kilian Q.},
  booktitle = {Advances in Neural Information Processing Systems},
  volume = {36},
  year = {2023},
  url = {https://proceedings.neurips.cc/paper_files/paper/2023/hash/b2a2bd5d5051ff6af52e1ef60aefd255-Abstract-Conference.html}
}

@article{ma2026dfm,
  title = {{Drift Flow Matching}},
  author = {Ma, Chenrui and Xiao, Xi and Zhao, Lin and Wang, Tianyang and Fioretto, Ferdinando and Shen, Yanning},
  journal = {arXiv preprint arXiv:2605.17244},
  year = {2026},
  url = {https://arxiv.org/abs/2605.17244}
}

@inproceedings{nie2025llada,
  title = {{Large Language Diffusion Models}},
  author = {Nie, Shen and Zhu, Fengqi and You, Zebin and Zhang, Xiaolu and Ou, Jingyang and Hu, Jun and Zhou, Jun and Lin, Yankai and Wen, Ji-Rong and Li, Chongxuan},
  booktitle = {Advances in Neural Information Processing Systems},
  year = {2025},
  volume = {38},
  url = {https://neurips.cc/virtual/2025/oral/130308}
}

@article{oba2026tokendrift,
  title = {{Drifting Objectives for Refining Discrete Diffusion Language Models}},
  author = {Oba, Daisuke and Furuta, Hiroki and Okazaki, Naoaki},
  journal = {arXiv preprint arXiv:2605.19470},
  year = {2026},
  url = {https://arxiv.org/abs/2605.19470}
}

@inproceedings{peebles2023dit,
  title = {{Scalable Diffusion Models with Transformers}},
  author = {Peebles, William and Xie, Saining},
  booktitle = {Proceedings of the IEEE/CVF International Conference on Computer Vision},
  year = {2023},
  pages = {4172--4182},
  doi = {10.1109/ICCV51070.2023.00387},
  url = {https://openaccess.thecvf.com/content/ICCV2023/html/Peebles_Scalable_Diffusion_Models_with_Transformers_ICCV_2023_paper.html}
}

@misc{qwen2026qwen35,
  title = {{Qwen3.5}: Towards Native Multimodal Agents},
  author = {{Qwen Team}},
  year = {2026},
  month = {feb},
  howpublished = {\url{https://qwen.ai/blog?id=qwen3.5}},
  note = {Model card of {Qwen3.5-9B}: \url{https://huggingface.co/Qwen/Qwen3.5-9B}}
}

@inproceedings{sahoo2024mdlm,
  title = {Simple and Effective Masked Diffusion Language Models},
  author = {Sahoo, Subham Sekhar and Arriola, Marianne and Schiff, Yair and Gokaslan, Aaron and Marroquin, Edgar and Chiu, Justin T. and Rush, Alexander and Kuleshov, Volodymyr},
  booktitle = {Advances in Neural Information Processing Systems},
  volume = {37},
  year = {2024},
  url = {https://proceedings.neurips.cc/paper_files/paper/2024/hash/eb0b13cc515724ab8015bc978fdde0ad-Abstract-Conference.html}
}

@InProceedings{sahoo2025duo,
  title = 	 {The Diffusion Duality},
  author =       {Sahoo, Subham Sekhar and Deschenaux, Justin and Gokaslan, Aaron and Wang, Guanghan and Chiu, Justin T and Kuleshov, Volodymyr},
  booktitle = 	 {Proceedings of the 42nd International Conference on Machine Learning},
  pages = 	 {52584--52619},
  year = 	 {2025},
  editor = 	 {Singh, Aarti and Fazel, Maryam and Hsu, Daniel and Lacoste-Julien, Simon and Berkenkamp, Felix and Maharaj, Tegan and Wagstaff, Kiri and Zhu, Jerry},
  volume = 	 {267},
  series = 	 {Proceedings of Machine Learning Research},
  month = 	 {13--19 Jul},
  publisher =    {PMLR},
  url = 	 {https://proceedings.mlr.press/v267/sahoo25a.html},
}

@inproceedings{savinov2022sundae,
  title = {{Step-unrolled Denoising Autoencoders for Text Generation}},
  author = {Savinov, Nikolay and Chung, Junyoung and Binkowski, Mikolaj and Elsen, Erich and van den Oord, Aaron},
  booktitle = {International Conference on Learning Representations},
  year = {2022},
  url = {https://iclr.cc/virtual/2022/poster/6942}
}

@article{shabalin2025tencdm,
  title = {{TEncDM: Understanding the Properties of the Diffusion Model in the Space of Language Model Encodings}},
  author = {Shabalin, Alexander and Meshchaninov, Viacheslav and Chimbulatov, Egor and Lapikov, Vladislav and Kim, Roman and Bartosh, Grigory and Molchanov, Dmitry and Markov, Sergey and Vetrov, Dmitry},
  journal = {Proceedings of the AAAI Conference on Artificial Intelligence},
  volume = {39},
  number = {23},
  pages = {25110--25118},
  year = {2025},
  doi = {10.1609/aaai.v39i23.34696},
  url = {https://ojs.aaai.org/index.php/AAAI/article/view/34696}
}

@inproceedings{snell2025scaling,
  title = {{Scaling LLM Test-Time Compute Optimally Can be More Effective than Scaling Parameters for Reasoning}},
  author = {Snell, Charlie and Lee, Jaehoon and Xu, Kelvin and Kumar, Aviral},
  booktitle = {International Conference on Learning Representations},
  year = {2025},
  url = {https://proceedings.iclr.cc/paper_files/paper/2025/hash/1b623663fd9b874366f3ce019fdfdd44-Abstract-Conference.html}
}

@inproceedings{song2023consistency,
  title = {{Consistency Models}},
  author = {Song, Yang and Dhariwal, Prafulla and Chen, Mark and Sutskever, Ilya},
  booktitle = {Proceedings of the 40th International Conference on Machine Learning},
  year = {2023},
  series = {Proceedings of Machine Learning Research},
  volume = {202},
  pages = {32211--32252},
  publisher = {PMLR},
  url = {https://proceedings.mlr.press/v202/song23a.html}
}

@article{su2021roformer,
  title = {{RoFormer}: Enhanced Transformer with Rotary Position Embedding},
  author = {Su, Jianlin and Lu, Yu and Pan, Shengfeng and Murtadha, Ahmed and Wen, Bo and Liu, Yunfeng},
  journal = {arXiv preprint arXiv:2104.09864},
  year = {2021},
  url = {https://arxiv.org/abs/2104.09864}
}

@inproceedings{wang2021codet5,
  title = {{C}ode{T}5: Identifier-aware Unified Pre-trained Encoder-Decoder Models for Code Understanding and Generation},
  author = {Wang, Yue  and Wang, Weishi  and Joty, Shafiq  and Hoi, Steven C.H.},
  booktitle = {Proceedings of the 2021 Conference on Empirical Methods in Natural Language Processing},
  month = {nov},
  year = {2021},
  address = {Online and Punta Cana, Dominican Republic},
  publisher = {Association for Computational Linguistics},
  url = {https://aclanthology.org/2021.emnlp-main.685/},
  doi = {10.18653/v1/2021.emnlp-main.685},
  pages = {8696--8708}
}

@inproceedings{wang2023selfconsistency,
  title = {{Self-Consistency Improves Chain of Thought Reasoning in Language Models}},
  author = {Wang, Xuezhi and Wei, Jason and Schuurmans, Dale and Le, Quoc and Chi, Ed and Narang, Sharan and Chowdhery, Aakanksha and Zhou, Denny},
  booktitle = {International Conference on Learning Representations},
  year = {2023},
  url = {https://openreview.net/forum?id=1PL1NIMMrw}
}

@article{wang2025hrm,
  title = {{Hierarchical Reasoning Model}},
  author = {Wang, Guan and Li, Jin and Sun, Yuhao and Chen, Xing and Liu, Changling and Wu, Yue and Lu, Meng and Song, Sen and Abbasi Yadkori, Yasin},
  journal = {arXiv preprint arXiv:2506.21734},
  year = {2025},
  url = {https://arxiv.org/abs/2506.21734}
}

@inproceedings{wang2025remdm,
  title = {{Remasking Discrete Diffusion Models with Inference-Time Scaling}},
  author = {Wang, Guanghan and Schiff, Yair and Sahoo, Subham Sekhar and Kuleshov, Volodymyr},
  booktitle = {Advances in Neural Information Processing Systems},
  year = {2025},
  volume = {38},
  url = {https://openreview.net/forum?id=IJryQAOy0p}
}

@article{yuan2023scaling,
  title = {{Scaling Relationship on Learning Mathematical Reasoning with Large Language Models}},
  author = {Yuan, Zheng and Yuan, Hongyi and Li, Chengpeng and Dong, Guanting and Lu, Keming and Tan, Chuanqi and Zhou, Chang and Zhou, Jingren},
  journal = {arXiv preprint arXiv:2308.01825},
  year = {2023},
  url = {https://arxiv.org/abs/2308.01825}
}

@inproceedings{zelikman2022star,
  title = {{STaR: Bootstrapping Reasoning With Reasoning}},
  author = {Zelikman, Eric and Wu, Yuhuai and Mu, Jesse and Goodman, Noah},
  booktitle = {Advances in Neural Information Processing Systems},
  volume = {35},
  pages = {15476--15488},
  year = {2022},
  url = {https://proceedings.neurips.cc/paper_files/paper/2022/hash/639a9a172c044fbb64175b5fad42e9a5-Abstract-Conference.html}
}

@inproceedings{
oba2026stopping,
title={Stopping Computation for Converged Tokens in Masked Diffusion-{LM} Decoding},
author={Daisuke Oba and Danushka Bollegala and Masahiro Kaneko and Naoaki Okazaki},
booktitle={The Fourteenth International Conference on Learning Representations},
year={2026},
url={https://openreview.net/forum?id=PzhNnMepgl}
}

\appendix

\section{Appendix}

\subsection{Encoding of the Prompt and the Target}\label{app:encoding}

The prompt and the target are encoded in two different ways.
The target latent is obtained by encoding the whole sequence, that is,
the problem statement followed by the solution, as one sequence.
With a contextual encoder, the latent at every solution position is thus
contextualized on the problem statement; with a per-token code such as the
one-hot code of the Sudoku experiments, the two encodings coincide position by
position.
The prompt latent \(P\) is obtained by encoding the problem statement alone.
The two tasks lay out the prompt and the solution differently.

\paragraph{GSM8K.}
We use the frozen encoder of CodeT5-small~\citep{wang2021codet5} (35.3M
parameters), whose contextualized embeddings have 512 dimensions, as the
T5-small encoder of ELF~\citep{hu2026elf} does.
We use CodeT5 because its tokenizer is designed for source code;
\citet[Section~4.2]{wang2021codet5} report that the default T5 tokenizer maps
code symbols such as \texttt{\{} and \texttt{\}} to unknown tokens.
In our own check of the T5-small tokenizer (\texttt{google-t5/t5-small}),
newlines and indentation were also lost and \texttt{\{}, \texttt{\}}, and
\texttt{<} became unknown tokens, whereas the CodeT5 tokenizer
(\texttt{Salesforce/codet5-small}) reproduced the same Python snippets
exactly.
Each example is tokenized with the CodeT5 tokenizer as BOS, the problem
statement, a newline, the program, and EOS, and padded to \(L=576\) tokens;
the prompt consists of BOS, the problem statement, and the newline.
The target latent is the output of the frozen CodeT5 encoder for the whole
sequence, in which the padding tokens are hidden from the attention keys but
still receive an output, so it covers all 576 positions, including the prompt
positions and the padding.
The prompt latent \(P\in\mathbb{R}^{576\times512}\) is the output of the
encoder for the prompt alone, placed at the prompt positions, with zeros at all
later positions.
The model outputs all 576 positions, but the noise and the draft enter the
model only at the positions after the prompt (the program, EOS, and the
padding), which are the positions where the prompt latent is zero, and the
drift field and the loss use only these positions; the outputs at the prompt
positions are never compared with a target and are discarded, so the next
round receives the draft only at the positions after the prompt
(Appendix~\ref{app:architecture}).
At test time, the problem statement is encoded in the same way, and the
generated program is read from the positions after the prompt up to the first
EOS.

Both latents are standardized with fixed scalar constants before they enter
the model:
\[
z \leftarrow \frac{z-\mu}{\sigma},
\qquad
\mu=0,\quad \sigma=0.3 .
\]
The same constants are used for the prompt latent, the target latent, and
the continuous output of the model.

\paragraph{Sudoku.}
The sequence is the 180-token format of S-FLM: BOS, the 89-token puzzle (81
cells, where an empty cell is the blank token, and 8 row separators), BOS, and
the 89-token answer, which is the solution grid with the same separators.
Each of the 12 tokens is encoded by a row of the \(12\times12\) identity matrix
(\(D_{\mathrm{emb}}=12\)), without standardization.
The model generates only the 89 answer positions (\(L=89\)): the target latent
is the code of the 89 answer tokens, and the prompt latent
\(P\in\mathbb{R}^{91\times12}\) is the code of the first 91 tokens.
Because the code of a token does not depend on its context, the puzzle does not
need to be part of the target.
The drift field and the loss use all 89 answer positions, including the row
separators.
All 89 answer tokens, including the given digits and the row separators, are
generated, and the metric is their exact match; the given digits are not
overwritten, and decoding is not restricted to digits.

We also experiment with context-free embedding tables on GSM8K (the CodeT5
input embedding table and a random Gaussian table); these results are reported
in Appendix~\ref{app:gsm8k_ablation}.

\subsection{Model Architecture and Input Layout}\label{app:architecture}

\paragraph{Input of the source formulation.}
Inside \(f_\theta\), the inputs are projected to the model dimension and
summed into the initial hidden representation
\[
h_r
=
\begin{cases}
\begin{aligned}[t]
&\epsilon_0 W_{\mathrm{noise}}
+
P W_{\mathrm{prompt}}\\
&\quad{}+
\mathbf{1}_L\phi(0)W_{\mathrm{round}},
\end{aligned}
& r=0, \\[3pt]
\begin{aligned}[t]
&d_r W_{\mathrm{draft}}
+
P W_{\mathrm{prompt}}\\
&\quad{}+
\mathbf{1}_L\phi(r)W_{\mathrm{round}},
\end{aligned}
& r\geq1.
\end{cases}
\]
The projection matrices
\(W_{\mathrm{noise}},W_{\mathrm{prompt}},W_{\mathrm{draft}}
\in\mathbb{R}^{D_{\mathrm{emb}}\times D_{\mathrm{model}}}\) and
\(W_{\mathrm{round}}\in\mathbb{R}^{10\times D_{\mathrm{model}}}\) have no bias,
and \(\mathbf{1}_L\) is a column vector of length \(L\) whose entries are all 1,
so that the round term is added at every position.
In the GSM8K experiments, \(P\), \(\epsilon_0\), and \(d_r\) all lie in
\(\mathbb{R}^{576\times512}\), and the terms occupy disjoint positions: before
the projection, the noise and the draft are multiplied by the indicator of the
positions where \(P\) is zero (the positions after the prompt), so the prompt
positions receive only the prompt latent and the later positions receive only
the noise or the draft; thus \(h_r\in\mathbb{R}^{576\times D_{\mathrm{model}}}\).
In the Sudoku experiments, the noise and the draft cover only the 89 answer
positions, so the prompt latent is placed before them instead of being added
to them:
\[
\begin{aligned}
h_r
&=
\operatorname{cat}_{\mathrm{seq}}
\bigl(
P W_{\mathrm{prompt}},\;
x_r
\bigr)
\\
&\quad{}+
\mathbf{1}_{180}\,\phi(r)W_{\mathrm{round}},
\\
x_r
&=
\begin{cases}
\epsilon_0 W_{\mathrm{noise}}, & r=0,\\
d_r W_{\mathrm{draft}}, & r\geq1,
\end{cases}
\end{aligned}
\]
with \(P\in\mathbb{R}^{91\times12}\) and
\(\epsilon_0,d_r\in\mathbb{R}^{89\times12}\), so that
\(h_r\in\mathbb{R}^{180\times D_{\mathrm{model}}}\); \(f_\theta\) returns the
outputs at the last 89 positions.

\paragraph{Round embedding.}
To provide the model with the current round, we use the round embedding
\(\phi(r)\), where \(r\) is the round index starting from 0.
Specifically, we use
\[
\begin{aligned}
\phi(r)
={}&
\bigl(
\mathbf{1}[r=0],\,
\frac{r}{1+r},
\\
&\quad
\sin(\pi r),\,
\sin(\pi r/2),
\\
&\quad
\sin(\pi r/4),\,
\sin(\pi r/8),
\\
&\quad
\cos(\pi r),\,
\cos(\pi r/2),
\\
&\quad
\cos(\pi r/4),\,
\cos(\pi r/8)
\bigr).
\end{aligned}
\]

\(\mathbf{1}[r=0]\) is 1 if and only if \(r=0\), and 0 otherwise.
Since \(r\) is an integer, \(\sin(\pi r)=0\) for every round; we keep this
constant coordinate so that the sine and cosine terms use the same four
frequencies.

\paragraph{Backbone.}
We feed \(h_r\) into \(g_\theta\), a stack of bidirectional
pre-normalization Transformer blocks followed by a final normalization layer.
No normalization is applied to \(h_r\) itself before the first block.
On both tasks, each block applies a scale-only LayerNorm before the attention
and before the feed-forward network, the feed-forward network is a GELU network,
and the final layer is a scale-only LayerNorm, following the compared methods;
on GSM8K, the blocks are the DiT blocks of the S-FLM implementation (those of
its autoregressive model, without adaLN), made bidirectional, with bias-free
attention projections and a feed-forward network of width 3,072 with biases
(Table~\ref{tab:gsm8k_settings}).
Positional information is provided by RoPE~\citep{su2021roformer} applied to
the queries and keys inside every attention layer of \(g_\theta\), as in the
backbones of the compared methods.
We do not add a positional embedding to \(h_r\).
No attention mask is used; on GSM8K, the padding positions take part in the
attention like the other positions.
The sizes of \(g_\theta\) are given in Appendix~\ref{app:settings}.

\paragraph{Output projection.}
Finally, we apply the output projection to the normalized output of
\(g_\theta\):
\[
f_\theta(\,\cdot\,,P,r)
=
g_\theta(h_r)
\left(
W_{\mathrm{out}}+W_r
\right)
+
b_0.
\]

Here,
\[
W_r
=
\begin{cases}
W_{\mathrm{init}}, & r=0,\\
W_{\mathrm{refine}}, & r\geq1.
\end{cases}
\]
Thus, we use different projections for the initial generation
at \(r=0\) and the refinement rounds at \(r\geq1\).
The input projections \(W_{\mathrm{noise}}\), \(W_{\mathrm{prompt}}\),
\(W_{\mathrm{draft}}\), and \(W_{\mathrm{round}}\) have no bias.
Among the output projections, only \(W_{\mathrm{out}}\) has the bias
\(b_0\); \(W_{\mathrm{init}}\) and \(W_{\mathrm{refine}}\) are bias-free.
The round index selects the input projection as it selects the output
projection.

\paragraph{Initialization.}
The round projection \(W_{\mathrm{round}}\) and the output projections
\(W_{\mathrm{init}}\) and \(W_{\mathrm{refine}}\) are initialized to zero, and
\(W_{\mathrm{out}}\) is drawn from \(\mathcal{N}(0,0.02^2)\) with a zero bias.
On GSM8K, \(W_{\mathrm{noise}}\) and \(W_{\mathrm{prompt}}\) are drawn
independently from \(\mathcal{N}(0,0.02^2)\) and \(W_{\mathrm{draft}}\) is
initialized to zero; on Sudoku, \(W_{\mathrm{noise}}\), \(W_{\mathrm{prompt}}\),
and \(W_{\mathrm{draft}}\) start from the same matrix drawn from
\(\mathcal{N}(0,0.02^2)\).
On both tasks, the Transformer blocks use the default initialization of
PyTorch.

\paragraph{Additive formulation.}
The projected draft is added to the projected noise,
\[
\begin{aligned}
h_r
&=
\epsilon_r W_{\mathrm{noise}}
+
P W_{\mathrm{prompt}}
\\
&\quad{}+
d_r W_{\mathrm{draft}}
+
\mathbf{1}_L \phi(r) W_{\mathrm{round}},
\end{aligned}
\]
with the same layout for each task as above (on Sudoku,
\(\epsilon_r W_{\mathrm{noise}}+d_r W_{\mathrm{draft}}\) occupies the 89 answer
positions after \(P W_{\mathrm{prompt}}\)); the rest of \(f_\theta\) is the
same as in the source formulation.
The initial draft \(d_0\) is the all-zero latent on GSM8K and the code of the
blank token at every answer position on Sudoku.

\paragraph{Conditional formulation.}
We concatenate the latent sequence \(h_r^{\mathrm{draft}}\), which represents
the draft from the previous round, to the latent sequence
\(h_r^{\mathrm{cur}}\), which is used for the current generation:
\[
\begin{aligned}
h_r^{\mathrm{cur}}
&=
\epsilon_r W_{\mathrm{noise}}
+
P W_{\mathrm{prompt}}
\\
&\quad{}+
\mathbf{1} \phi(r) W_{\mathrm{round}}
+
\mathbf{1} e_{\mathrm{cur}},
\\
h_r^{\mathrm{draft}}
&=
d_r W_{\mathrm{draft}}
+
\mathbf{1} \phi(r) W_{\mathrm{round}}
+
\mathbf{1} e_{\mathrm{draft}},
\end{aligned}
\]
where \(h_r^{\mathrm{cur}}\) has the layout of the source formulation for each
task, \(e_{\mathrm{cur}}\) and \(e_{\mathrm{draft}}\) are learnable embeddings
that distinguish the two sequences, and \(\mathbf{1}\) is a column vector of
ones of the corresponding length.
The two sequences are concatenated along the sequence dimension,
\(H_r=\operatorname{cat}_{\mathrm{seq}}(h_r^{\mathrm{cur}},h_r^{\mathrm{draft}})\),
and fed into \(g_\theta\).
Each draft position receives the same RoPE position as the corresponding
position of the current generation.
The next draft is obtained by applying the output projection of the source
formulation to the hidden states of \(g_\theta(H_r)\) at the generated
positions of \(h_r^{\mathrm{cur}}\).
We report the conditional formulation only on Sudoku.
There, \(h_r^{\mathrm{cur}}\) consists of the 91 prompt positions followed by
the 89 answer positions, which carry \(\epsilon_r W_{\mathrm{noise}}\), and
\(e_{\mathrm{cur}}\) is added to all 180 positions; \(h_r^{\mathrm{draft}}\)
has the 89 answer positions, which receive the RoPE positions of the answer
positions of \(h_r^{\mathrm{cur}}\).
Attention therefore runs over 269 positions instead of the 180 positions of the
source and additive formulations.

\subsection{Drift Field and Loss}\label{app:drift_field}

This section gives the exact form of the drift field and the loss of
Section~\ref{sec:training_additive}.
Let \(x_1,\ldots,x_n\in\mathbb{R}^{L\times D_{\mathrm{emb}}}\) be the generated
samples of one problem at the terminal round, and let the positives be encoded as
in Appendix~\ref{app:encoding}; with a single positive, we write it as \(y\).
Norms of latent sequences are taken over all their entries.

\paragraph{Scaled coordinates.}
Following the official implementation of the Drifting Model,
the drift field is computed in scaled coordinates, and the repulsive term
is evaluated at several kernel temperatures rather than with a single
bandwidth \(\tau\).
Let \(G\) denote the set of coordinates at the positions that enter the drift
field and the loss, that is, the positions after the prompt (all 89 answer
positions on Sudoku; Appendix~\ref{app:encoding}); let \(\|\cdot\|_G\) denote
the Euclidean norm restricted to these coordinates, and let \(|G|\) be their
number, averaged over the mini-batch.
Let \(z_1,\ldots,z_{n+1}\) be the \(n\) generated samples followed by the
target, that is, \(z_k=x_k\) for \(k\leq n\) and \(z_{n+1}=y\),
and let
\[
\begin{gathered}
\Delta_{jk}
=
\|x_j-z_k\|_G,
\\
j=1,\ldots,n,\;
k=1,\ldots,n+1,
\end{gathered}
\]
be the distances from each generated sample to every reference,
including the self distance \(\Delta_{jj}\).
Let \(m\) be the mean of all \(\Delta_{jk}\) over the mini-batch,
and define the scale
\[
\begin{gathered}
\kappa
=
\max
\left(
\frac{m}{\sqrt{|G|}},
\;\delta
\right),
\\
\tilde{x}_j=\frac{x_j}{\kappa},
\qquad
\tilde{y}=\frac{y}{\kappa},
\end{gathered}
\]
where \(\delta=10^{-3}\) prevents division by a vanishing scale.
The scale \(\kappa\) and every quantity below except \(\tilde{x}_j\) are
treated as constants (stop-gradient).

\paragraph{Attractive drift.}
In general, the attractive drift is the kernel-weighted barycenter of the
targets minus the sample.
When \(N_{+}\) targets (positives) \(y_1,\ldots,y_{N_{+}}\) are available for
the same prompt,
\[
\begin{aligned}
V^{+}_j
&=
\sum_{i=1}^{N_{+}}
w^{+}_{ji}\,\tilde{y}_i
-
\tilde{x}_j,
\\
w^{+}_{ji}
&=
\frac{
\exp\left(-\|x_j-y_i\|_G/\tau^{+}\right)
}{
\displaystyle
\sum_{i'=1}^{N_{+}}
\exp\left(-\|x_j-y_{i'}\|_G/\tau^{+}\right)
}.
\end{aligned}
\]
With several positives, the references \(z_k\) of the scaled coordinates
(and of the repulsive affinity below) are the \(n\) generated samples followed
by all \(N_{+}\) positives.
Unlike the original Drifting Model, which normalizes one kernel matrix over the
positives and the generated samples together, we normalize the attractive
weights over the positives only.
With the joint normalization, the attractive weights of the positives of one
problem became almost uniform in our experiments, so the attraction pointed to
the centroid of the positives, which rarely decodes to a valid program.
The repulsive term keeps the joint normalization of the official
implementation and is then restricted to the generated samples (see below).
The bandwidth \(\tau^{+}\) is \(0.01\) times the median distance between
two distinct positives of the same prompt over the mini-batch; pairs whose
distance is at most \(10^{-4}\) (identical programs) are excluded from the
median.
We use the median of the positive set rather than a fixed value because the
positive queue (Section~\ref{sec:verifier}) adds positives during training
and the spread of the positive set changes; the factor \(0.01\) was chosen
empirically (Appendix~\ref{app:gsm8k_ablation} reports the factors 0.1 and 1).
With such a small bandwidth, the attractive weight concentrates on the
nearest positive.
In the TinyGSM experiments, \(N_{+}\) is the number of positives in the queue of
the prompt, between \(K\) and \(K+4\) (Section~\ref{sec:verifier}).
In the Sudoku experiments, the answer is unique and only one target \(y\) is
available for each prompt (\(N_{+}=1\)).
Then the softmax has a single entry, \(w^{+}_{j1}=1\), and the attractive
drift reduces to the difference between the target and the sample,
\(V^{+}_j=\tilde{y}-\tilde{x}_j\).

\paragraph{Repulsive drift with multiple temperatures.}
The repulsive drift uses the kernel weights of the official implementation,
which are computed at three temperatures.
Let \(\bar{\Delta}_{jk}=\Delta_{jk}/m\) be the normalized distances, where a large
constant is added to the self entries \(\bar{\Delta}_{jj}\) so that a sample does
not repel itself.
For each temperature \(\tau\in\{0.02,\,0.05,\,0.2\}\), we form the logits
\(\ell_{jk}=-\bar{\Delta}_{jk}/\tau\) and the doubly normalized affinity
\[
a^{(\tau)}_{jk}
=
\sqrt{
\operatorname{softmax}_{k}(\ell_{j\cdot})_{k}
\;
\operatorname{softmax}_{j}(\ell_{\cdot k})_{j}
},
\]
where the first softmax is over the \(n+N_+\) references of sample \(j\) and
the second is over the \(n\) generated samples of reference \(k\).
Restricting to the other generated samples and renormalizing gives the
weights
\[
w^{(\tau)}_{jk}
=
\frac{a^{(\tau)}_{jk}}{\sum_{\ell\leq n,\,\ell\neq j}a^{(\tau)}_{j\ell}},
\qquad
k\leq n,\; k\neq j,
\]
and the repulsive drift at temperature \(\tau\) is
\[
V^{-,(\tau)}_j
=
\sum_{k\neq j}
w^{(\tau)}_{jk}
\left(
\tilde{x}_k-\tilde{x}_j
\right).
\]
A small temperature makes the weights concentrate on the nearest generated
sample, while a large temperature spreads them over all generated samples;
using the three temperatures together captures both the local and the
global structure of the generated samples.

\paragraph{Drift field and loss.}
For each temperature, the drift field combines the two terms with the
coefficients \(\lambda^{+}\) and \(\lambda^{-}\),
\[
V^{(\tau)}_j
=
\lambda^{+}V^{+}_j-\lambda^{-}V^{-,(\tau)}_j .
\]
The original Drifting Model uses \((\lambda^{+},\lambda^{-})=(1,1)\)
(\(V^{+}-V^{-}\)); our experiments use \((2,1)\) (\(2V^{+}-V^{-}\)) unless noted
otherwise (Section~\ref{sec:experiments}), and Appendix~\ref{app:lambda}
compares the two.
Let \(\rho_\tau\) be the root mean square of \(V^{(\tau)}\) over all generated
samples in the mini-batch and all coordinates in \(G\).
The final drift is the sum of the normalized fields over the three
temperatures,
\[
V_j
=
\sum_{\tau\in\{0.02,\,0.05,\,0.2\}}
\frac{V^{(\tau)}_j}{\rho_\tau}.
\]
We then construct the stop-gradient target as
\[
x^{\mathrm{tgt}}_j
=
\operatorname{sg}
\left(
\tilde{x}_j+V_j
\right).
\]
Appendix~\ref{app:lambda} works out this target for a single positive.

For one training example, the loss is
\[
\mathcal{L}(\theta)
=
\frac{1}{n L D_{\mathrm{emb}}}
\sum_{j=1}^{n}
\left\|
\tilde{x}_j-x^{\mathrm{tgt}}_j
\right\|_G^2 ,
\]
where the gradient flows only through \(\tilde{x}_j=x_j/\operatorname{sg}(\kappa)\).

The loss is applied only at the sampled terminal round.
Appendix~\ref{app:training_procedure} describes the full procedure over a
mini-batch step by step and summarizes it in
Algorithm~\ref{alg:iterative_training}.

\subsection{Coefficients of the Attractive and Repulsive Drifts}\label{app:lambda}

We write the drift field as \(\lambda^{+}V^{+}-\lambda^{-}V^{-}\)
(Section~\ref{sec:experiments}; Appendix~\ref{app:drift_field} gives the exact
form with the three temperatures).
The original Drifting Model uses \((\lambda^{+},\lambda^{-})=(1,1)\), and we
use \((2,1)\) in all experiments except the \((1,1)\) ablations.
This appendix recalls why \((1,1)\) is the natural choice of the original
model, compares the two choices on Sudoku and on GSM8K, and gives a possible
reason for the difference.

\paragraph{The original Drifting Model.}
With \((1,1)\), the field \(V^{+}-V^{-}\) is anti-symmetric,
\(V_{p,q}=-V_{q,p}\), so it vanishes when the generated distribution matches
the target (Section~\ref{sec:drifting}), and the original paper reports that
the non-anti-symmetric field \(2V^{+}-V^{-}\) makes training fail on
ImageNet~\citep{deng2026generativemodelingdrifting}.
The original model also trains with many generated samples per class, so the
kernel weights of the repulsion depend on the distances and a sample is pushed
mainly by the generated samples near it.
In our setting, the field with \((2,1)\) is not anti-symmetric, and neither is
our attraction, which is normalized over the positives only
(Appendix~\ref{app:drift_field}); matching the target distribution is
therefore no longer guaranteed to be an equilibrium.
With a single positive, all generated samples coinciding with it still give
\(V_j=0\) for any coefficients, although this does not guarantee a fixed point
of the recurrent generation map.

\paragraph{Sudoku.}
We selected \((\lambda^{+},\lambda^{-})=(2,1)\) using a separately
obtained Sudoku development dataset. Neither the S-FLM benchmark
evaluation scores nor the GSM8K test scores were used to select
these coefficients. We use the same coefficient pair on both tasks
and \(n=2\) generated samples to limit training cost: in the source
formulation, \(n=2\) approximately halves the training cost relative
to \(n=4\) (Appendix~\ref{app:training_cost}).

Table~\ref{tab:sudoku_lambda} reports benchmark evaluation ablations
of the coefficients and the number of generated samples.
With \(n=2\), \((1,1)\) collapses to 0.0\% at every evaluated NFE
for all three seeds, whereas \((2,1)\) reaches 92.1\% at NFE=16.
With \(n=4\), both choices train, and \((1,1)\) achieves higher
accuracy from NFE=8 onward.

\begin{table*}[tp]
\centering
\small
\resizebox{\textwidth}{!}{%
\begin{tabular}{lrrrrrrrrr}
\hline
Setting & NFE=1 & 4 & 5 & 8 & 16 & 32 & 64 & 128 & 180 \\
\hline
\((2,1)\), \(n=2\) (default) & 7.3\(\pm\)1.0 & 78.9\(\pm\)0.6 & 82.5\(\pm\)0.8 & 88.4\(\pm\)0.4 & 92.1\(\pm\)0.4 & 93.1\(\pm\)0.6 & 93.5\(\pm\)0.7 & 93.9\(\pm\)0.7 & 94.0\(\pm\)0.7 \\
\((2,1)\), \(n=4\) & 7.6\(\pm\)2.2 & 78.1\(\pm\)1.4 & 81.7\(\pm\)1.3 & 87.4\(\pm\)1.5 & 91.1\(\pm\)1.2 & 92.1\(\pm\)1.1 & 92.6\(\pm\)1.2 & 92.7\(\pm\)1.2 & 92.9\(\pm\)1.4 \\
\((1,1)\), \(n=2\) & 0.0\(\pm\)0.0 & 0.0\(\pm\)0.0 & 0.0\(\pm\)0.0 & 0.0\(\pm\)0.0 & 0.0\(\pm\)0.0 & 0.0\(\pm\)0.0 & 0.0\(\pm\)0.0 & 0.0\(\pm\)0.0 & 0.0\(\pm\)0.0 \\
\((1,1)\), \(n=4\) & 6.5\(\pm\)1.7 & 76.0\(\pm\)2.6 & 80.7\(\pm\)2.2 & 88.0\(\pm\)1.2 & 93.5\(\pm\)0.6 & 95.9\(\pm\)0.4 & 97.1\(\pm\)0.3 & 98.0\(\pm\)0.2 & 98.3\(\pm\)0.2 \\
\hline
\end{tabular}}
\caption{Sudoku Hard accuracy (\%) at 20,000 updates of the source formulation with the drift-field coefficients \((\lambda^{+},\lambda^{-})\in\{(2,1),(1,1)\}\) and \(n\in\{2,4\}\) generated samples (2,000 validation puzzles, one candidate per puzzle, EMA weights, mean \(\pm\) standard deviation over three seeds, round embedding clamped to \(\phi(\min(r,4))\); the other settings as in Section~\ref{sec:sudoku}). The first row is the ReLaDrift Source row of Table~\ref{tab:sudoku_results}.}
\label{tab:sudoku_lambda}
\end{table*}

\paragraph{GSM8K.}
On GSM8K (\(K=12\), the verifier, and the positive queue), \((1,1)\) collapses
with \(n=2\) and with \(n=4\): the accuracy is 0.0\% at every NFE at the
evaluations at 20,000 and 40,000 updates, after which both runs were stopped,
whereas \((2,1)\) with \(n=2\) reaches 16.1\% at NFE=16 at 40,000 updates and
29.4\% at 250,000 updates (Tables~\ref{tab:gsm8k_ablation}
and~\ref{tab:gsm8k_matched}).
Unlike on Sudoku, four generated samples do not rescue \((1,1)\) here.

\paragraph{A possible reason: the restoring force toward the positives.}
Let \(\tilde{x}_j=x_j/\kappa\) and \(\tilde{y}_i=y_i/\kappa\)
denote the scaled generated samples and positives, respectively,
where \(\kappa\) is the common mini-batch scale defined in
Appendix~\ref{app:drift_field}.
The barycenters below are computed in these scaled coordinates.

Write the attractive and repulsive drifts of a generated sample \(x_j\) as
\(V^{+}_j=\bar{y}^{+}_j-\tilde{x}_j\) and
\(V^{-}_j=\bar{y}^{-}_j-\tilde{x}_j\), where \(\bar{y}^{+}_j\) is the
kernel-weighted barycenter of the positives and \(\bar{y}^{-}_j\) that of the
other generated samples (Appendix~\ref{app:drift_field}); both sets of weights
sum to one.
Ignoring the RMS normalization for the moment, the stop-gradient target of
\(x_j\) is
\[
\begin{aligned}
&\tilde{x}_j+\lambda^{+}V^{+}_j-\lambda^{-}V^{-}_j\\
&\quad=
\bigl(1-\lambda^{+}+\lambda^{-}\bigr)\tilde{x}_j
+\lambda^{+}\bar{y}^{+}_j-\lambda^{-}\bar{y}^{-}_j .
\end{aligned}
\]
With \((1,1)\), the coefficient of \(\tilde{x}_j\) is exactly 1, whatever the
kernel weights, the number of generated samples, and the number of positives:
the displacement asked of \(x_j\), \(\bar{y}^{+}_j-\bar{y}^{-}_j\), does not
depend on the position of \(x_j\) itself except through the kernel weights.
A common displacement of all generated samples of a problem away from the
positives is corrected, because it moves \(\bar{y}^{-}_j\) but not
\(\bar{y}^{+}_j\); this is the mechanism of the original model, where the
field acts on the generated distribution as a whole.
A sample that strays alone, however, while the other generated samples stay
near the positives, is not pulled back: when \(\bar{y}^{-}_j=\bar{y}^{+}_j\),
the field on \(x_j\) vanishes wherever \(x_j\) is.
With \((2,1)\), the coefficient of \(\tilde{x}_j\) is 0: the target is
\(\bar{y}^{+}_j+(\bar{y}^{+}_j-\bar{y}^{-}_j)\), anchored at the barycenter of
the positives, so each sample is pulled toward the positives individually, in
addition to the repulsion.
With the RMS normalization of Appendix~\ref{app:drift_field}, the same
computation gives the coefficient \(1-\gamma(\lambda^{+}-\lambda^{-})\) with
\(\gamma=\sum_{\tau}1/\rho_\tau\): again 1 for \((1,1)\), and \(1-\gamma\)
for \((2,1)\).

The case \(n=2\) makes the consequence visible.
Each sample \(j\) then has a single negative, the other generated sample
\(k\), and the normalization over the other generated samples degenerates for
every temperature \(\tau\):
\[
w^{(\tau)}_{jk}
=
\frac{a^{(\tau)}_{jk}}{a^{(\tau)}_{jk}}
=
1,
\qquad
V^{-,(\tau)}_j
=
\tilde{x}_k-\tilde{x}_j ,
\]
whatever the distance between the two samples, so the repulsion grows linearly
with their distance, the opposite of the local repulsion that the kernel is
meant to give.
With a single positive, \(V^{+}_j=\tilde{y}-\tilde{x}_j\), and the field
\[
V^{(\tau)}_j
=
\lambda^{+}(\tilde{y}-\tilde{x}_j)-\lambda^{-}(\tilde{x}_k-\tilde{x}_j)
\]
is the same at every temperature, so \(V_j=\gamma\,V^{(\tau)}_j\).
For the targets \(x^{\mathrm{tgt}}_j=\operatorname{sg}(\tilde{x}_j+V_j)\) and
\(x^{\mathrm{tgt}}_k=\operatorname{sg}(\tilde{x}_k+V_k)\) of the two samples,
this gives
\[
x^{\mathrm{tgt}}_j-x^{\mathrm{tgt}}_k
=
\bigl(1+\gamma\,(2\lambda^{-}-\lambda^{+})\bigr)\,(\tilde{x}_j-\tilde{x}_k).
\]
With \((1,1)\), the factor is \(1+\gamma>1\): every update asks the two
samples to move further apart, nothing bounds their separation, and training
collapses, as observed on both tasks.
With \((2,1)\), the factor is exactly 1: the target keeps the current
separation, and the restoring force above pulls both samples toward the
positive.
On GSM8K, the attraction concentrates on the nearest positive
(Appendix~\ref{app:drift_field}), so the same relation holds approximately
when the two samples share their nearest positive.
With \(n=4\), the repulsive weights depend on the distances again and the
separation no longer grows mechanically, which is consistent with \((1,1)\)
training on Sudoku, where every generated sample is attracted to the single
positive.
However, even with \(n=4\), the coefficient of \(\tilde{x}_j\) in the
stop-gradient target remains 1 under \((1,1)\): the explicit
\(\tilde{x}_j\) terms in the attractive and repulsive drifts cancel.
The drift therefore still lacks an explicit restoring term proportional
to \(-\tilde{x}_j\), although the kernel weights depend on the sample's
position.
We conjecture that this absence contributes to the collapse of \((1,1)\)
on GSM8K even with \(n=4\), where generated samples can be attracted to
different positives among the \(K=12\) given ones and the queue.
With \((2,1)\), an explicit restoring term remains.

\subsection{Training Procedure of the Recurrent Model}\label{app:training_procedure}

This section describes one optimizer step of the source formulation
(Section~\ref{sec:training_additive}) over a mini-batch of \(B\) problems in
words; Algorithm~\ref{alg:iterative_training} summarizes it.
The additive and conditional formulations differ in the recurrence, which
draws fresh noise at every round and starts from the initial draft \(d_0\), and
in where the generated samples branch (Section~\ref{sec:training_additive}),
and the changes for the verifier and the positive queue are given in
Section~\ref{sec:verifier} and Appendix~\ref{app:verified_step}.

\paragraph{Step 1: sampling the number of rounds.}
The number of rounds \(R_b\sim\operatorname{Uniform}\{1,\ldots,R_{\max}\}\) is
drawn for each problem \(b\); the terminal round of problem \(b\) is round
\(R_b-1\).
On GSM8K, the draw is independent for every problem, and intermediate round
\(r\) is one batched forward pass over the trajectories whose \(R_b>r+1\); on
Sudoku, one \(R\) is drawn per mini-batch and shared by all \(B\) problems, so
that every intermediate round is one batched forward pass over the
mini-batch.

\paragraph{Step 2: encoding the prompt and the positives.}
The prompt latent \(P_b\) and the positives \(y_{b,i}\) are obtained as in
Appendix~\ref{app:encoding}; on GSM8K, both are standardized with the fixed
constants \((\mu,\sigma)\).
The encoder is frozen, so these latents are computed once and cached; in the
TinyGSM experiments the latents of the \(K\) given positives are precomputed and
only the programs added to the positive queue are encoded during training.

\paragraph{Step 3: intermediate rounds without gradient.}
For each problem \(b\), \(n\) independent noises
\(\epsilon'_{b,j}\sim\mathcal{N}(0,I)\) are drawn at all \(L\) positions, one
per trajectory, and the rounds \(r=0,\ldots,R_b-2\) of each trajectory of
problem \(b\) are computed under stop-gradient: the first round transports the noise,
\(d_1=\operatorname{sg}(f_\theta(\epsilon'_{b,j},P_b,0))\), and each later round
is applied to the previous output with the round embedding \(\phi(r)\),
\(d_{r+1}=\operatorname{sg}(f_\theta(d_r,P_b,r))\).
No new noise is drawn after the first round, no decoding or re-encoding takes
place between rounds, and no activations are stored, so these passes cost only
forward computation.
In the additive and conditional formulations, one intermediate trajectory is
started from the fixed initial draft \(d_0\) with fresh noise at every round,
shared by the generated samples, and the branching happens at the terminal
round.
When \(R_b=1\), this step is empty for problem \(b\), whose terminal round is
then the first round.

\paragraph{Step 4: terminal round with gradient.}
At round \(R_b-1\), each trajectory of problem \(b\) gives its generated
sample \(x_{b,j}=f_\theta(d_{R_b-1},P_b,R_b-1)\) (or
\(x_{b,j}=f_\theta(\epsilon'_{b,j},P_b,0)\) when \(R_b=1\)), computed with
gradient.
With the verifier, each \(x_{b,j}\) is decoded and executed at this point, and
verified programs are pushed into the positive queue before the drift is
computed.

\paragraph{Steps 5--9: drift field, loss, and update.}
The scale \(\kappa\), the attractive and repulsive drifts, the RMS-normalized
field with \((\lambda^{+},\lambda^{-})=(2,1)\), the stop-gradient targets
\(x^{\mathrm{tgt}}_{b,j}\), and the loss are computed from the terminal-round
samples \(x_{b,j}\) and the positives of each problem exactly as in
Appendix~\ref{app:drift_field}; with the verifier, the repulsive weights of
verified samples are set to zero (Appendix~\ref{app:verified_step}).
The gradient flows only through the terminal-round outputs
\(\tilde{x}_{b,j}=x_{b,j}/\operatorname{sg}(\kappa)\); the intermediate rounds
and every other quantity are constants.
The gradient norm is clipped to 1, one AdamW step is taken, and the EMA weights
are updated.

\begin{algorithm}[htbp]
\caption{One training step of ReLaDrift (source formulation)}
\label{alg:iterative_training}
\begin{algorithmic}[1]
\Require Mini-batch of \(B\) problems with prompts \(P_b\) and positives \(\{y_{b,i}\}\), model \(f_\theta\), maximum number of rounds \(R_{\max}\), number of generated samples \(n\), drift-field coefficients \((\lambda^{+},\lambda^{-})\), temperatures \(\mathcal{T}=\{0.02,0.05,0.2\}\)
\State Sample the number of rounds \(R_b\sim\operatorname{Uniform}\{1,\ldots,R_{\max}\}\) for each problem \(b\) (one shared \(R\) per mini-batch on Sudoku)
\For{\(b=1,\ldots,B\) and \(j=1,\ldots,n\)}
    \State Sample \(\epsilon'_{b,j}\sim\mathcal{N}(0,I)\) and set \(d\gets\epsilon'_{b,j}\) \Comment{one trajectory per generated sample}
    \For{\(r=0,\ldots,R_b-2\)} \Comment{intermediate rounds, no gradient}
        \State \(d\gets\operatorname{sg}\bigl(f_\theta(d,P_b,r)\bigr)\) \Comment{round 0 transports the noise; later rounds refine the previous output}
    \EndFor
    \State \(x_{b,j}\gets f_\theta(d,P_b,R_b-1)\) \Comment{terminal round, with gradient}
\EndFor
\State Compute the mini-batch scale \(\kappa\) and the scaled coordinates \(\tilde{x}_{b,j}\gets x_{b,j}/\kappa\), \(\tilde{y}_{b,i}\gets y_{b,i}/\kappa\) (Appendix~\ref{app:drift_field})
\For{each generated sample \((b,j)\)}
    \State \(V^{+}_{b,j}\gets\) kernel-weighted barycenter of the positives of problem \(b\) minus \(\tilde{x}_{b,j}\)
    \State \(V^{-,(\tau)}_{b,j}\gets\) affinity-weighted repulsion from the other generated samples of problem \(b\), for each \(\tau\in\mathcal{T}\)
    \State \(V^{(\tau)}_{b,j}\gets\lambda^{+}V^{+}_{b,j}-\lambda^{-}V^{-,(\tau)}_{b,j}\)
\EndFor
\State \(V_{b,j}\gets\sum_{\tau\in\mathcal{T}}V^{(\tau)}_{b,j}/\rho_\tau\), where \(\rho_\tau\) is the root mean square of \(V^{(\tau)}\) over the mini-batch
\State \(x^{\mathrm{tgt}}_{b,j}\gets\operatorname{sg}\bigl(\tilde{x}_{b,j}+V_{b,j}\bigr)\)
\State \(\mathcal{L}(\theta)\gets\frac{1}{nBLD_{\mathrm{emb}}}\sum_{b,j}\|\tilde{x}_{b,j}-x^{\mathrm{tgt}}_{b,j}\|_G^2\), and update \(\theta\) with its gradient, which flows only through \(\tilde{x}_{b,j}\)
\end{algorithmic}
\end{algorithm}

\subsection{Training Step with the Verifier and the Positive Queue}\label{app:verified_step}

Algorithm~\ref{alg:verified_step} shows one optimizer step of the source
formulation with the verifier and the positive queue (Section~\ref{sec:verifier}).
It extends Algorithm~\ref{alg:iterative_training} by decoding and executing the
terminal samples, updating the positive queues, and removing the repulsion from
verified samples.

Learning from generated reasoning selected using known answers has precedents
in STaR~\citep{zelikman2022star} and rejection sampling
fine-tuning~\citep{yuan2023scaling}; here, accepted programs enter the drifting
positive set, and their corresponding generated samples are removed from the
negative set.
\citet[Appendix~A.8]{deng2026generativemodelingdrifting} cache real samples in
queues; our per-problem queue stores generated programs accepted by an
execution-based check.

\paragraph{Pseudo-reference.}
The verifier uses a voted pseudo-reference from the numerical answers of the
\(K\) given positives, not a ground-truth answer.
The vote selects the unique largest answer group (a plurality, without
requiring more than half the votes).
Values are sorted and grouped when their distance from the group's minimum
is at most \(10^{-3}\); the reference is the value from the earliest selected
positive in the winning group.
If the largest group is tied, or if the tolerance relation is nontransitive,
the verifier is disabled for that problem, which is then trained with its
\(K\) given positives only; no generated sample is accepted or masked.
The original TinyGSM solution is not used to construct this reference.
Self-consistency aggregates sampled reasoning paths to select an answer at
inference time~\citep{wang2023selfconsistency}; here, voting supplies a
training-time pseudo-reference.
Agreement with it does not establish ground-truth correctness.

\paragraph{Execution.}
At the sampled terminal round, each generated sample is decoded into tokens with
the decoder (Appendix~\ref{app:decoder}), the program after the prompt (up to
the first EOS) is executed in a sandbox with a time limit of 15~ms, and the
sample is verified if its return value matches the reference answer within
\(10^{-3}\).

\paragraph{Positive queue.}
A verified program is written back as the prompt followed by the program and
EOS and is encoded with the same encoder in the same way as the given positives
(Appendix~\ref{app:encoding}).
It is pushed into the four additional slots only if it is not already in the
queue, and only the programs added during training are pushed out; the \(K\)
given positives are never pushed out, even when the queue is full.
A verified program enters the queue before the loss of the same step is
computed.

\paragraph{Repulsion mask.}
With the weights \(w^{(\tau)}_{jk}\) of Appendix~\ref{app:drift_field}, the
repulsive drift becomes
\[
V^{-,(\tau)}_j
=
\sum_{k\neq j}
w^{(\tau)}_{jk}\,(1-v_k)
\left(
\tilde{x}_k-\tilde{x}_j
\right),
\]
that is, the weights of verified samples are set to zero after the
normalization, without renormalizing the remaining weights.
The root mean square \(\rho_\tau\) is computed from the field before this
masking.
The loss itself is applied to all generated samples, including the verified
ones.

\paragraph{Sudoku.}
On Sudoku, the verifier checks the rules of Sudoku and the agreement with the
clues.
Because the answer is unique, a verified answer is always the given positive
and the queue never grows (we observed \(N_{+}=1\) throughout training); there
the verifier acts only through the repulsion mask.

\begin{algorithm}[htbp]
\caption{One training step of the source formulation with the verifier and the positive queue}
\label{alg:verified_step}
\begin{algorithmic}[1]
\Require Problems \(\{c_b\}_{b=1}^{B}\), model \(f_\theta\), decoder,
positive queues \(\{\mathcal{Q}_c\}\), pseudo-references \(\{a_c\}\) (votes
of the \(K\) given positives), number of generated samples \(n\),
maximum number of rounds \(R_{\max}\), drift-field coefficients \((\lambda^{+},\lambda^{-})\)
\State Encode the prompts \(P_b\) (zeros after the prompt) and set \(G_b\) to
the positions after the prompt of problem \(c_b\)
\State Sample the number of rounds \(R_b\sim\operatorname{Uniform}\{1,\ldots,R_{\max}\}\) for each problem \(b\)
\For{\(b=1,\ldots,B\) and \(j=1,\ldots,n\)}
    \State Sample \(\epsilon_{b,j}\sim\mathcal{N}(0,I)\) and run the source recurrence
    for \(R_b\) rounds; rounds \(0,\ldots,R_b-2\) are computed without gradient
    \State \(x_{b,j}\gets\) output of the terminal round \(R_b-1\) (with gradient)
\EndFor
\For{\(b=1,\ldots,B\) and \(j=1,\ldots,n\)}
    \State Decode \(x_{b,j}\), execute the program, and set
    \(v_{b,j}\gets\mathbf{1}[|\text{output}-a_{c_b}|\leq10^{-3}]\) (\(v_{b,j}\gets0\) if execution fails or \(c_b\) has no valid pseudo-reference)
    \If{\(v_{b,j}=1\)}
        \State Push the program into \(\mathcal{Q}_{c_b}\) (the \(K\) given positives are never pushed out)
    \EndIf
\EndFor
\State Encode the programs newly pushed into \(\mathcal{Q}_{c_b}\) (the latents of the \(K\) given positives are cached) and collect the positives \(y_{b,1},\ldots,y_{b,N_{+,b}}\) of \(\mathcal{Q}_{c_b}\)
\State Compute the drift \(V_{b,j}\) with these positives, the coefficients
\((\lambda^{+},\lambda^{-})\), and the repulsion mask \(v\), and set \(x^{\mathrm{tgt}}_{b,j}\gets\operatorname{sg}(\tilde{x}_{b,j}+V_{b,j})\)
(Appendix~\ref{app:drift_field} and this section)
\State \(\mathcal{L}\gets\frac{1}{nBLD_{\mathrm{emb}}}\sum_{b,j}\|\tilde{x}_{b,j}-x^{\mathrm{tgt}}_{b,j}\|_{G_b}^2\);
backpropagate, clip the gradient norm to 1, and take an AdamW step
\State Update the EMA weights
\end{algorithmic}
\end{algorithm}

\subsection{Generation Procedure}\label{app:generation_procedure}

Algorithm~\ref{alg:generation} gives the generation procedure of
Section~\ref{sec:generation}: the prompt is encoded once, the draft is updated
\(N\) times with the same model \(f_\theta\), with the round embedding clamped
beyond the training depth, and the decoder is applied once to the final draft.

\begin{algorithm}[htbp]
\caption[Generation with \(N\) function evaluations]{Generation with \(N\) function evaluations.\newline \textcolor{red}{Red: source formulation;} \textcolor{blue}{blue: additive and conditional formulations.}}
\label{alg:generation}
\begin{algorithmic}[1]
\Require Problem statement, model \(f_\theta\), decoder, number of rounds \(N\) (the NFE), training depth \(R_{\max}\); initial draft \(d_0\) (additive and conditional formulations only)
\Ensure Generated token sequence
\State \(P \gets\) prompt latent of the problem statement (Section~\ref{sec:embeddings})
\State Sample \(\epsilon\sim\mathcal{N}(0,I)\)
\If{source formulation}
    \State \textcolor{red}{\(d \gets f_\theta(\epsilon, P, 0)\)} \Comment{first round transports the noise}
\Else
    \State \textcolor{blue}{\(d \gets f_\theta(\epsilon, d_0, P, 0)\)} \Comment{first round: noise and the initial draft}
\EndIf
\For{\(r=1,\ldots,N-1\)}
    \State \(\bar{r} \gets \min(r, R_{\max}-1)\) \Comment{clamp the round embedding beyond the training depth}
    \If{source formulation}
        \State \textcolor{red}{\(d \gets f_\theta(d, P, \bar{r})\)} \Comment{refine the previous output; no new noise}
    \Else
        \State Sample \(\epsilon\sim\mathcal{N}(0,I)\)
        \State \textcolor{blue}{\(d \gets f_\theta(\epsilon, d, P, \bar{r})\)} \Comment{fresh noise; the previous output is the draft}
    \EndIf
\EndFor
\State \Return the tokens decoded from \(d\) (Section~\ref{sec:embeddings})
\end{algorithmic}
\end{algorithm}

\subsection{Decoder}\label{app:decoder}

This appendix describes the learned decoder used with the pretrained contextual
encoder (CodeT5) in the GSM8K experiments; on Sudoku, decoding is the inner
product with the one-hot codes followed by the argmax (Section~\ref{sec:sudoku}).
The decoder is a separate bidirectional Transformer (8 layers, width 512,
8 heads; 42.2M parameters) that takes a sequence of latents as input and
predicts a token distribution at every position, as in TEncDM~\citep{shabalin2025tencdm}.
It is trained before the Drifting Model and kept frozen during its training and
at test time.
It is trained to recover the token sequence from perturbed latents with the
corruption that ELF uses to train decoding~\citep[Appendix~C.1]{hu2026elf}: at every
position, the clean latent \(z\) is replaced by \(\alpha z+(1-\alpha)\epsilon\)
with \(\epsilon\sim\mathcal{N}(0,I)\) and \(\alpha=\operatorname{sigmoid}(0.8+0.8\,\xi)\),
\(\xi\sim\mathcal{N}(0,1)\), drawn independently for each position.
The training data are the CodeT5 latents of one Qwen3.5 solution for each of
100,000 training problems, and the validation data are 10,000 held-out
TinyGSM problems with their original solutions; we train for 20,000 updates
with batch size 64, AdamW (learning rate \(3\times10^{-4}\) after 500 warm-up
updates, decayed to \(3\times10^{-5}\), weight decay 0.01), and use the
checkpoint with the lowest validation loss, at 18,000 updates.
The decoder is trained on latents standardized with its own statistics
\((\mu_{\mathrm{dec}},\sigma_{\mathrm{dec}})\), which differ from the
constants \((\mu,\sigma)\) used by the Drifting Model
(Appendix~\ref{app:encoding}).
Therefore, a generated latent \(z\) is re-standardized by the affine map
\[
z_{\mathrm{dec}}
=
\frac{z\,\sigma+\mu-\mu_{\mathrm{dec}}}{\sigma_{\mathrm{dec}}}
\]
before it is given to the decoder.
The same decoder is used by the verifier during training (Section~\ref{sec:verifier}).
On the clean CodeT5 latents of 10,000 training problems that were not used to
train the decoder (one Qwen3.5 solution each), the decoder recovers 99.98\% of
the tokens and 98.18\% of the programs exactly, and 99.55\% of the decoded
programs return the value of the original program; of the 182 programs that
are not recovered exactly, 140 differ only in the docstring or by a consistent
renaming of a variable and return the same value.
These numbers concern clean latents, not the latents generated by ReLaDrift.

\subsection{Generating the Positives}\label{app:qwen_prompt}

We use the \texttt{train} split of \texttt{TinyGSM/TinyGSM}, revision
\texttt{5049a7840fa546985eaae4c698d5daf6fdf7b47d}, containing 11,846,109
question--program pairs.
This is the size of the released dataset used here; the original
paper~\citep{liu2023tinygsm} reports 12.3M pairs.
We retain the 11,811,180 pairs that fit within 576 CodeT5 tokens, then use
\texttt{datasets.Dataset.train\_test\_split} with a validation fraction of
0.01 and seed 42 to obtain 11,693,068 training pairs and 118,112 validation
pairs.
We generate alternatives for the first 2,483,048 problems of this shuffled
training split (21.0\% of the unfiltered release).

We sampled 32 Python solutions per problem using
\texttt{Qwen/Qwen3.5-9B}~\citep{qwen2026qwen35} at revision
\texttt{c202236235762e1c871ad0ccb60c8ee5ba337b9a}, with the
thinking mode off, temperature 0.85, top-p 0.95, top-k 20, min-p 0, no
presence, frequency, or repetition penalty, and at most 600 generated tokens,
using the prompt below, which we wrote for this work; the required output
format (one function \texttt{simple\_math\_problem} whose docstring repeats
the problem statement) follows the TinyGSM solutions.
A solution is kept if it is executable, which we check with a restricted
interpreter, fits within 576 tokens of the CodeT5 tokenizer used by ReLaDrift
and 512 tokens of the SmolLM tokenizer used by the compared methods, and differs
from the other kept solutions of the same problem after removing comments,
docstrings, and whitespace.
In the kept solutions, we remove the comments but keep the docstrings.
The training set consists of the 1,588,781 problems (13.4\% of TinyGSM) with at
least 12 kept solutions, with 12 solutions each; all 12 return the same value
for 62.7\% of the problems.
The compared methods are trained on all 12 solutions of these problems, expanded into 19,065,372 shuffled problem--solution pairs; ReLaDrift uses the first \(K\) solutions of each problem as its given positives.
Except for the comments, the generated programs are similar in style to the
TinyGSM solutions.
Neither the generated nor the original solutions are guaranteed to be correct,
and we do not filter them by correctness: for one problem in our data, the
TinyGSM solution and two of the generated solutions return three different
values.
In the prompt, \texttt{\{\{ question \}\}} is replaced by the problem statement; its line breaks are adjusted to the column width here.

\begingroup\footnotesize
\begin{verbatim}
You are given a grade-school math word problem.
Generate a correct Python solution that
satisfies
all of the following requirements:
1. Define exactly one function with this
   signature:

def simple_math_problem():

2. Immediately after the function definition,
   include a triple-quoted docstring containing
   the given problem statement exactly as
   provided.
3. Solve the problem using ordinary Python
   arithmetic.
4. Use meaningful variables for quantities
   appearing in the problem.
5. Solve the problem step by step using
   intermediate variables when natural.
6. The function must return the final numerical
   answer.
7. The program must be executable and must
   return the mathematically correct answer.
8. Do not print the answer.
9. Do not use Markdown code fences.
10. Do not include any explanation or text
    outside the Python function.
11. Keep the solution natural and similar in
    style to real TinyGSM solutions. Do not use
    unnecessarily sophisticated Python
    features, helper functions, classes, loops,
    comprehensions, symbolic algebra, or
    external libraries unless they are
    genuinely necessary for the problem.
12. Keep comments minimal. Do not perform
    calculations in comments, and do not
    include numerical values in comments.
13. The numerical computation must be derived
    from the quantities in the problem. Do not
    simply return the known final answer
    directly.

### Question

{{ question }}

### Program
\end{verbatim}
\endgroup

\subsection{Experimental Settings}\label{app:settings}

Tables~\ref{tab:sudoku_settings} and~\ref{tab:gsm8k_settings} summarize the settings
of ReLaDrift and the compared methods on Sudoku (Section~\ref{sec:sudoku}) and
GSM8K (Section~\ref{sec:gsm8k}).
On both tasks, AR, MDLM, Duo, and S-FLM are built from the S-FLM repository
(\url{https://github.com/jdeschena/s-flm}, commit \texttt{30def79}) and FMLM+
from the FMLM+ repository
(\url{https://github.com/MananAg007/posterior-refinement}, commit
\texttt{73652f9}); neither repository is modified.
On GSM8K, we run the training pipelines of the two repositories.
On Sudoku, all methods share our single-GPU training loop, which calls the
model, loss, optimizer, learning-rate schedule, and EMA of each repository, so
that all methods see the same puzzles in the same order: the 48,000 training
puzzles are permuted anew at every epoch and the incomplete last batch of the
permutation (128 puzzles, different ones in each epoch) is dropped, so an epoch
consists of 187 batches of 256 puzzles.
For FRM, we use our own implementation following its
paper~\citep{helbling2026frm}.
On Sudoku, the S-FLM setting is the one of the official Sudoku script of S-FLM
(the DiT on the sphere and the truncated, adaptive noise schedule), and the
S-FLM paper also uses the standard DiT on Sudoku.
On GSM8K, it is the one of the official TinyGSM script, which follows the
TinyGSM experiments of the S-FLM paper (the S-arch backbone, 163.0M
parameters), run
without the run-time patch that we use on Sudoku (Table~\ref{tab:sudoku_settings});
with the official data loader, the adaptive schedule is first refitted at
update 37,251.

On GSM8K, the compared methods pad the sequences to 512 tokens and compute the
loss on the padding, as in S-FLM.
On GSM8K, a generated program is correct if its return value matches the
answer exactly for integers and within \(10^{-3}\) otherwise, with a time limit
of 5~s.
ReLaDrift uses the optimization settings of the compared methods: AdamW
with \(\beta=(0.9,0.999)\), no weight decay, gradient clipping at 1, dropout
0.1, and the warm-up-corrected EMA decay \(\min(0.9999,(u+1)/(u+10))\) at
update \(u\) of the official implementation.
In ReLaDrift, the dropout (on the attention and feed-forward outputs) is active
only in the terminal round, which receives the gradient; the earlier rounds of
an update are computed in evaluation mode.

\paragraph{Sudoku benchmark.}
The 2,000 validation puzzles of the benchmark of S-FLM are used only for
evaluation; we verified that each of them has a unique solution and that they
share no solution grid with the training puzzles.
Of its three difficulty levels (40, 35, and 30 of the 81 digits visible), we use the hard level, which has lower reported accuracy than the easier levels~\citep[Table~1]{deschenaux2026sflm}.
The Sudoku-Extreme benchmark~\citep{wang2025hrm} is nearly saturated
(99.5\% for FRM;~\citealp{helbling2026frm}).

\paragraph{Drift Flow Matching.}
DFM~\citep{ma2026dfm} trains a mean-velocity network \(u_\theta(x_t,t,r)\) on
the linear interpolant \(x_t=t\,x_1+(1-t)\,x_0\), \(x_0\sim\mathcal{N}(0,I)\),
with the drifting objective applied to the predictions
\(\hat{x}_r=x_t+(r-t)\,u_\theta\), so that a sample can be generated in one
step (\(t=0\), \(r=1\)) or refined with more Euler steps.
Its official code (commit \texttt{b8f287c}) covers class-conditional image
latents only, so we implement the prompt-conditioned case following
Appendix~D of the paper: for each puzzle, the solution \(x_1\) is paired with
\(n_g=2\) source draws \(x_{0,i}\), the positives are
\(x_{r,i}=r\,x_1+(1-r)\,x_{0,i}\), and the negatives are the predictions
\(\hat{x}_{r,i}\) of the model from the same \(x_{0,i}\).
The drift field, the law of the time pairs, the loss, and the sampler are
those of the official code: Sinkhorn-normalized Gibbs weights with the
Euclidean distance and temperature 1 (20 iterations; the negatives include
the sample itself), four time pairs per update with \(t=\min\) and
\(r=\max\) of \(\sigma(2.5z-1)\) and \(\sigma(2.5z'+1)\),
\(z,z'\sim\mathcal{N}(0,1)\), the squared error to the stopped target
\(\hat{x}_r+V\) summed over the coordinates, and Euler steps on the uniform
grid of \([0,1]\).
The network is the DiT of FMLM+ (Table~\ref{tab:sudoku_settings}; 28.61M
parameters; \(t\) and \(r-t\) enter through its two time embeddings), and the
data, batching, optimizer, EMA, budget, and evaluation are those of the other
methods.
Because every puzzle has a single solution, the Sinkhorn weights are
numerically the identity for \(r\lesssim0.7\), so on Sudoku the DFM objective
reduces to a regression toward the paired target \(x_{r,i}\); the group size
then changes mainly the number of source draws per puzzle, and \(n_g=2\)
matches the two trajectories of ReLaDrift.
With \(n_g\in\{4,8,16\}\), trained in the same way, the accuracy at NFE=16 is
31.9, 35.2, and 33.4\% (mean over three seeds), 3--6 points above \(n_g=2\)
(29.2\%) and within the seed standard deviation (5--8 points); the ranking
against ReLaDrift is unchanged.
No Sudoku result of DFM has been published.

\paragraph{Accuracy of our runs of the compared methods.}
On Sudoku, our runs of S-FLM and MDLM are within 2.4 points of the published
accuracies and our Duo runs are 3.1 points above them, whereas our FMLM+ runs
are about 4 points below the published results: at NFE=180, the sampling budget
of \citet[Table~1]{deschenaux2026sflm}, our S-FLM, Duo, and MDLM reach 43.2, 61.5,
and 32.6\%, compared with the 45.0, 58.4, and 30.2\% reported there.
Our FMLM+ reaches 67.0 and 77.5\% at NFE=4 and 16, compared with 71.2 and
81.4\% reported by \citet[Appendix~D, Table~13]{agarwal2026posterior}.
Among the compared parallel generators, FMLM+ is the strongest at every NFE from
4 to 128, and FRM with the SDE sampler is the strongest at NFE=180 (79.4\% versus
77.2\% for FMLM+); Duo reaches 61.9\% at NFE=16, and S-FLM, MDLM, and DFM
reach at most 43.2\%.
For FRM, the reported checkpoint at 20,000 updates is the end of Stage B, and
it is less accurate than the Stage A checkpoint of the same runs at 10,000
updates (Euler sampler, NFE=16: 71.8\% versus 86.1\%, mean over three seeds).
\citet{helbling2026frm} train on 1,000 Sudoku-Extreme puzzles with symmetry
augmentation, which we do not use; on the 48,000 puzzles of our setting, Stage
A already fits the training puzzles (its training loss is below 0.03 at 10,000
updates), so that the rollouts of Stage B contain few errors to correct.
We nevertheless report the checkpoint at 20,000 updates, which matches the
training budget of the other methods.
The autoregressive model reaches 0.5\% with greedy decoding and with sampling
(Table~\ref{tab:sudoku_results}), in agreement with the 1.0\% (greedy) and
0.6\% (sampling) reported on the hard level
by \citet[Table~1]{deschenaux2026sflm}.
On GSM8K, at NFE=1,024, our S-FLM, Duo, and MDLM runs are comparable to the
values reported at \(T=0.1\) with the original TinyGSM solutions
(about 18, 36, and 33\%; \citealp[Section~4.2 and Figure~3]{deschenaux2026sflm}).

\paragraph{Artifacts and licenses.}
TinyGSM~\citep{liu2023tinygsm} (Hugging Face dataset \texttt{TinyGSM/TinyGSM}) and GSM8K~\citep{cobbe2021gsm8k} (Hugging Face dataset \texttt{openai/gsm8k}) are released under the MIT License.
Qwen3.5-9B~\citep{qwen2026qwen35} and the SmolLM-135M
tokenizer~\citep{allal2024smollm} are released under the Apache License 2.0.
The CodeT5 reference implementation~\citep{wang2021codet5} is released
under the BSD 3-Clause License.
The Hugging Face model card for the \texttt{Salesforce/codet5-small}
checkpoint lists Apache License 2.0.\footnote{\url{https://huggingface.co/Salesforce/codet5-small}}
The S-FLM repository~\citep{deschenaux2026sflm} states the Apache License 2.0 in its README (the repository contains no license file), the FMLM+ repository~\citep{agarwal2026posterior} is released under the Apache License 2.0, and the Sudoku generator~\citep{alp2024sudoku} states the MIT License in its README.
The reference implementation of the Drifting Model~\citep{deng2026generativemodelingdrifting}
at commit \texttt{c8b4fee}, whose drift field we follow
(Appendix~\ref{app:drift_field}), did not include a license file.
The current official repository states that its code and pretrained
weights are released under the MIT License.\footnote{\url{https://github.com/lambertae/drifting}}
Our implementation is written from scratch.
All artifacts are used for research purposes in accordance with these terms. Our code is released under the Apache License 2.0.

\begin{table*}[tp]
\centering
\footnotesize
\begin{tabular}{p{1.6cm}p{2.9cm}p{1.9cm}p{2.0cm}p{2.0cm}p{2.0cm}}
\hline
Item & ReLaDrift (ours) & S-FLM & FMLM+ & FRM (our implementation) & DFM (our implementation) \\
\hline
Sequence, vocabulary & 180 tokens, vocabulary 12, fixed prefix 91 & same & same & same & same \\
State & continuous vector (12-dim one-hot code) & learned embedding on the sphere & one-hot + Gaussian noise & one-hot + carry & one-hot + Gaussian noise (linear interpolant \(x_t=t\,x_1+(1-t)\,x_0\)) \\
Encoder / decoder & one-hot code / inner product with the codes + argmax & learned embedding on the sphere / \(\arg\max\) of the output layer & one-hot / argmax & one-hot / argmax & one-hot / argmax \\
Transformer & 9 layers, width 512, 8 heads, FFN 2,048, RoPE, dropout 0.1, round embedding \(\phi(r)\) added to the input & DiT on the sphere, 8 layers, width 512, 8 heads & DiT, 8 layers, width 512, 8 heads, two-time adaLN & DiT, 8 layers, width 512, 8 heads, adaLN & DiT of FMLM+, 8 layers, width 512, 8 heads, two-time adaLN (\(t\), \(r-t\)) \\
Parameters & 28.39M & 28.56M & 28.61M & 28.57M & 28.61M \\
Optimizer & AdamW 3e-4 (warmup 2,500, then constant), $\beta=(0.9, 0.999)$, weight decay 0, clip 1 & same (official) & same (official) & same; optimizer and EMA restarted at Stage B & same \\
EMA & decay \(\min(0.9999,(u+1)/(u+10))\) & same & same & same & same \\
Training budget & 20,000 updates, batch 256 & 20,000 updates, batch 256 & 20,000 updates, batch 256 & Stage A 10,000 + Stage B 10,000, 256 puzzles $\times$ 4 states & 20,000 updates, batch 256 \\
Forward passes in training & gradient only at the terminal round; \(n=2\) independent trajectories & 1 & 2 (self-distillation) & FPF (rollout of depth 16) & 2 (\(n_g=2\) source draws per puzzle) \\
Generation & recurrent, NFE 1--180, \(\phi(\min(r,4))\) & ODE with the exact velocity, \(\arg\max\) at the last step, NFE 1--180 & Posterior Refinement, threshold 0.999, 1--180 rounds & Euler / SDE / held-time, NFE 1--180 & Euler on the uniform grid, NFE 1--180 \\
Scoring & exact match on the 89 answer tokens, no overwriting & same & same & same & same \\
\hline
\end{tabular}
\caption{Sudoku settings of ReLaDrift and the compared methods. AR, Duo, and MDLM are trained with the implementations in the S-FLM repository and its settings (the same 8-layer, width-512 DiT~\citep{peebles2023dit}, 28.6M parameters; the AR model is causal and has no adaLN, 25.2M). DFM (Drift Flow Matching) is our implementation following its paper and official code (commit \texttt{b8f287c}): the drift field, the law of the time pairs, the loss, and the Euler sampler are those of the official code, and the solution of each puzzle is paired with \(n_g=2\) source draws as in Appendix~D of its paper. Duo and MDLM use ancestral sampling at temperature 1, and AR is evaluated with greedy decoding and with sampling at temperature 1. S-FLM uses its truncated, adaptive noise schedule; all methods drop the last, incomplete batch of each epoch (128 of the 48,000 puzzles; the permutation is redrawn at every epoch, so the dropped puzzles differ), whereas the official loaders keep it; with this batching, the official code never refits the adaptive schedule, so we added a patch at run time, without modifying the repository, so that the schedule is refitted (first refit at update 1,051). The single-round Drifting Model uses the ReLaDrift column without the round embedding, the draft projection, and the refinement output projection (28.37M parameters) and runs one round. All methods use dropout 0.1 and bf16 mixed precision.}
\label{tab:sudoku_settings}
\end{table*}

\begin{table*}[tp]
\centering
\footnotesize
\begin{tabular}{p{1.7cm}p{3.3cm}p{2.1cm}p{2.0cm}p{2.3cm}}
\hline
Item & ReLaDrift (ours) & AR & S-FLM & FMLM+ \\
\hline
Positives & first \(K\) of the 12 solutions (fixed) + up to 4 verified samples (positive queue) & 12 (problem, solution) pairs per problem, shuffled & same & same \\
Tokenizer, length & CodeT5 (vocabulary 32,100), 576 & SmolLM-135M (vocabulary 49,153), 512 & same & same \\
Latent & CodeT5 contextualized embedding, 512 dim (frozen) & discrete tokens & learned embedding on the sphere & one-hot + Gaussian noise \\
Transformer & DiT blocks of S-FLM (bidirectional, no adaLN), 12 layers, width 768, 12 heads, FFN 3,072, RoPE, round embedding \(\phi(r)\) & DiT, 12 layers, width 768, 12 heads (causal) & S-arch, 12 layers, width 768, 12 heads & DiT, 12 layers, width 768, 12 heads, two-time adaLN \\
Parameters & core 87.4M (+ frozen encoder 35.3M, decoder 42.2M = 164.9M) & 160.5M & 163.0M & 168.0M \\
Optimizer & AdamW $\beta=(0.9, 0.999)$, weight decay 0, clip 1 (same as the compared methods) & AdamW $\beta=(0.9, 0.999)$, weight decay 0 (= Adam), clip 1 (official) & same & same \\
Dropout & 0.1 (terminal round only) & 0.1 & 0.1 & 0.1 \\
Objective & Source, \((\lambda^{+},\lambda^{-})=(2,1)\), temperatures 0.02/0.05/0.2, \(\tau^{+}=0.01\times\)median, verifier and positive queue & next-token cross-entropy & official & official \\
Verifier in training & yes (execution of the 2 terminal samples) & no & no & no \\
Forward passes in training & 2 trajectories of $R \sim U\{1..5\}$ rounds per problem, stop-gradient on intermediate rounds; decoding and execution of the 2 terminal samples & 1 & 1 & 2 (self-distillation) \\
Generation (official setting) & recurrent, NFE 1--1,024, one trajectory & sequential sampling (temperature 1, top-p 1; we report \(T=0.1\)), up to 512 tokens including the prompt & 1,024 ODE steps, exact velocity & Posterior Refinement, 32 rounds (NFE 32), threshold 0.999 \\
Decoding & 42.2M decoder, argmax per position & -- & argmax & argmax \\
\hline
\end{tabular}
\caption{GSM8K settings. All methods use the same 1,588,781 problems, batch size 512, learning rate 3e-4 after 2,500 warm-up updates (then constant), 250,000 updates, dropout 0.1, and the EMA decay \(\min(0.9999,(u+1)/(u+10))\) of the official S-FLM implementation. Duo (167.9M) and MDLM (167.9M) are trained with the same code base and settings as S-FLM; S-FLM, Duo, and MDLM sample with 1,024 steps in the S-FLM paper~\citep{deschenaux2026sflm}. All methods use bf16 mixed precision. The single-round Drifting Model uses the ReLaDrift column without the round embedding, the draft projection, and the refinement output projection (core 86.2M parameters). Only ReLaDrift and the single-round Drifting Model use a verifier during training.}
\label{tab:gsm8k_settings}
\end{table*}

\subsection{Training and Generation Cost}\label{app:training_cost}

\paragraph{Training.}
ReLaDrift runs several forward passes of \(f_\theta\) per update, whereas AR, S-FLM, Duo, and MDLM run one pass, FMLM+ about two (self-distillation), and DFM two (\(n_g=2\) source draws per puzzle); FRM propagates four states per puzzle and, in Stage B, additionally runs a rollout without gradient for fixed-point forcing.
With \(R\sim\operatorname{Uniform}\{1,\ldots,5\}\), the expected depth is \(\bar{R}=3\).
The source formulation propagates \(n=2\) independent trajectories, so it runs \(2\bar{R}=6\) forward passes per update, of which only the two terminal passes carry gradients.
The additive formulation shares one intermediate trajectory and branches into two candidates at the terminal round, so it runs \(\bar{R}-1+2=4\) forward passes, again with two gradient passes.
The conditional formulation runs the same passes as the additive one, but on Sudoku its attention covers 269 instead of 180 positions, because the 89 draft tokens are appended.
Counting a backward pass as twice a forward pass, one update of the source formulation costs about \((4+2\times3)/3\approx3.3\) times an update of a one-pass method, and the additive formulation about \((2+2\times3)/3\approx2.7\) times; the intermediate passes run without gradient and store no activations.
Measured on one H200 for Sudoku (batch 256 puzzles, \(n=2\)), one update takes 0.22~s for Source, so 20,000 updates take about 1.2~h; Additive takes 0.17~s per update (0.9~h), Conditional 0.26~s (1.4~h), and the single-round Drifting Model 0.12~s (0.7~h).
With \(n=4\), Source takes 0.42~s per update (2.3~h).
On GSM8K, the verifier additionally decodes the two terminal samples with the decoder, executes the programs (time limit 15~ms each), and encodes the new positives of the queue with the CodeT5 encoder; these steps run without gradient.
On 8 H200 GPUs, one update including these steps takes 0.56, 0.65, and 0.76~s for \(K=1\), 4, and 12, so 250,000 updates take about 39, 45, and 53~h.
The decoder (Appendix~\ref{app:decoder}) is trained once, before ReLaDrift, for 20,000 updates on one GPU in less than one hour, which is less than 0.3\% of the GPU time of training ReLaDrift with \(K=12\) (8 GPUs for about 53~h).

\paragraph{Generation.}
Generation costs NFE forward passes of \(f_\theta\) and at most one pass each of
the encoder and the decoder, which do not depend on NFE.
On GSM8K, the frozen CodeT5 encoder (35.3M parameters) encodes the prompt once
per problem, which can be cached, and the decoder (42.2M parameters) decodes the
final draft; both are smaller than the 87.4M-parameter \(f_\theta\).
On Sudoku, the codes are fixed and decoding is an \(\arg\max\) over 12 codes, so
no extra pass is needed.
On one H200, evaluating all 2,000 Sudoku puzzles at every NFE up to 16 (one
shared 16-round trajectory, 16 passes per puzzle) takes about 3~s.
On GSM8K, with batches of 64 problems on one H200, a round of \(f_\theta\) takes 24~ms, a step of Duo 95~ms, a step of MDLM 93~ms, a refinement round of FMLM+ 130~ms, and a step of S-FLM 242~ms (its sampler computes the velocity in double precision); generating the 1,319 test programs takes ReLaDrift 6.8~s at NFE=4 and 13~s at NFE=16, Duo 63~s at NFE=32, and S-FLM 5,200~s at NFE=1,024 (Figure~\ref{fig:walltime}).

\subsection{Additional Sudoku Results}\label{app:sudoku_full}

Table~\ref{tab:sudoku_full} lists all eleven ReLaDrift conditions of the unified
Sudoku Hard study (three seeds each, 20,000 updates, EMA, clamped round
embedding) at every evaluated NFE, including the single-round Drifting
Model, the source formulation with the verifier, and the training
depths \(R_{\max}\in\{3,8,16\}\).
On Sudoku, the verifier can only remove the repulsion from verified samples,
because the answer is unique and the queue never grows; with \(n=2\) it makes
little difference: it is 1.1--1.7 points more accurate at NFE=2--5 and within
0.3 points at NFE=1 and NFE \(\geq 8\) (92.0\% versus 92.1\% at NFE=16).
The training depths are discussed in Section~\ref{sec:sudoku}; this table adds
NFE=2.

\begin{table*}[tp]
\centering
\scriptsize
\setlength{\tabcolsep}{4pt}
\begin{tabular}{lrrrrrrrrrr}
\hline
Condition & NFE=1 & 2 & 4 & 5 & 8 & 16 & 32 & 64 & 128 & 180 \\
\hline
Source (main) & 7.3\(\pm\)1.0 & 54.7\(\pm\)0.3 & 78.9\(\pm\)0.6 & 82.5\(\pm\)0.8 & 88.4\(\pm\)0.4 & 92.1\(\pm\)0.4 & 93.1\(\pm\)0.6 & 93.5\(\pm\)0.7 & 93.9\(\pm\)0.7 & 94.0\(\pm\)0.7 \\
Additive & 7.1\(\pm\)2.2 & 54.8\(\pm\)3.9 & 77.6\(\pm\)2.6 & 81.5\(\pm\)3.1 & 86.2\(\pm\)3.6 & 89.5\(\pm\)3.1 & 90.5\(\pm\)3.0 & 91.0\(\pm\)2.8 & 91.6\(\pm\)2.5 & 91.6\(\pm\)2.8 \\
Conditional & 7.4\(\pm\)0.9 & 54.3\(\pm\)3.3 & 77.6\(\pm\)1.7 & 81.1\(\pm\)1.1 & 86.1\(\pm\)1.2 & 89.6\(\pm\)1.8 & 90.7\(\pm\)1.7 & 91.2\(\pm\)2.0 & 91.6\(\pm\)2.3 & 91.7\(\pm\)2.2 \\
Single-round Drifting Model & 5.1\(\pm\)0.8 & 4.9\(\pm\)0.8 & 4.9\(\pm\)0.8 & 5.0\(\pm\)0.8 & 4.9\(\pm\)0.8 & 4.9\(\pm\)0.8 & 4.9\(\pm\)0.8 & 4.9\(\pm\)0.8 & 4.9\(\pm\)0.8 & 4.9\(\pm\)0.8 \\
Source, w/ verifier & 7.0\(\pm\)1.0 & 56.4\(\pm\)0.6 & 80.0\(\pm\)1.0 & 83.7\(\pm\)0.7 & 88.4\(\pm\)0.7 & 92.0\(\pm\)0.6 & 93.0\(\pm\)0.6 & 93.4\(\pm\)0.8 & 93.6\(\pm\)0.8 & 93.7\(\pm\)0.8 \\
Source, \((1,1)\) & 0.0\(\pm\)0.0 & 0.0\(\pm\)0.0 & 0.0\(\pm\)0.0 & 0.0\(\pm\)0.0 & 0.0\(\pm\)0.0 & 0.0\(\pm\)0.0 & 0.0\(\pm\)0.0 & 0.0\(\pm\)0.0 & 0.0\(\pm\)0.0 & 0.0\(\pm\)0.0 \\
Source, \((1,1)\), \(n=4\) & 6.5\(\pm\)1.7 & 50.7\(\pm\)4.2 & 76.0\(\pm\)2.6 & 80.7\(\pm\)2.2 & 88.0\(\pm\)1.2 & 93.5\(\pm\)0.6 & 95.9\(\pm\)0.4 & 97.1\(\pm\)0.3 & 98.0\(\pm\)0.2 & 98.3\(\pm\)0.2 \\
Source, \(n=4\) & 7.6\(\pm\)2.2 & 55.1\(\pm\)3.0 & 78.1\(\pm\)1.4 & 81.7\(\pm\)1.3 & 87.4\(\pm\)1.5 & 91.1\(\pm\)1.2 & 92.1\(\pm\)1.1 & 92.6\(\pm\)1.2 & 92.7\(\pm\)1.2 & 92.9\(\pm\)1.4 \\
Source, \(R_{\max}=3\) & 5.7\(\pm\)2.7 & 45.5\(\pm\)5.8 & 66.2\(\pm\)3.4 & 68.5\(\pm\)2.9 & 71.4\(\pm\)2.1 & 72.6\(\pm\)2.0 & 72.9\(\pm\)1.9 & 73.1\(\pm\)1.9 & 73.1\(\pm\)2.0 & 73.2\(\pm\)2.0 \\
Source, \(R_{\max}=8\) & 6.4\(\pm\)2.0 & 59.1\(\pm\)4.3 & 83.4\(\pm\)1.7 & 86.9\(\pm\)1.7 & 91.2\(\pm\)1.7 & 94.9\(\pm\)1.3 & 96.0\(\pm\)1.0 & 96.6\(\pm\)1.0 & 96.8\(\pm\)0.9 & 96.9\(\pm\)0.8 \\
Source, \(R_{\max}=16\) & 4.2\(\pm\)2.1 & 56.3\(\pm\)6.1 & 85.2\(\pm\)2.9 & 88.8\(\pm\)2.1 & 93.0\(\pm\)1.7 & 96.1\(\pm\)0.8 & 97.2\(\pm\)0.5 & 97.4\(\pm\)0.4 & 97.5\(\pm\)0.4 & 97.6\(\pm\)0.4 \\
\hline
\end{tabular}
\caption{Sudoku Hard accuracy (\%) of all ReLaDrift conditions at 20,000 updates (2,000 puzzles, one candidate, EMA, mean \(\pm\) standard deviation over three seeds, round embedding clamped to \(\phi(\min(r,R_{\max}-1))\)). Unless noted: source formulation, \((\lambda^{+},\lambda^{-})=(2,1)\), \(n=2\), \(R_{\max}=5\), no verifier. At NFE \(\geq 2\), the single-round Drifting Model is iterated, with its decoded output fed back in place of the noise (the dashed line of Figure~\ref{fig:teaser}; Appendix~\ref{app:naive_recurrence}).}
\label{tab:sudoku_full}
\end{table*}

\subsection{Results without the Clamp of the Round Embedding}\label{app:clamp}

The model uses the round embedding clamped to \(\phi(\min(r,R_{\max}-1))\), so
that it never receives a round embedding that it did not see in training
(Section~\ref{sec:generation}).
This appendix reports, on the same checkpoints, the accuracy when \(\phi(r)\) is
instead extrapolated to the rounds \(r\geq R_{\max}\); for NFE \(\leq R_{\max}\)
the two are identical (Tables~\ref{tab:sudoku_clamp} and~\ref{tab:gsm8k_clamp}).
The clamp only keeps the round embedding within the values seen in training;
it does not make the drafts after more than \(R_{\max}\) rounds follow the
distribution of the drafts seen in training.

On Sudoku, some runs collapse without the clamp.
For the source formulation, one seed falls to 3.0\% at NFE=8 and to at most
0.1\% at NFE \(\geq 16\); of the other two, one falls to 32.5\% at NFE=8 and
the other to 59.3\% at NFE=16, whereas with the clamp all three seeds stay at
88--93\% at NFE=8 and 16.
The additive and conditional formulations also have seeds that fall at NFE=8
or 16.
With \(R_{\max}=3\), all three seeds fall to 0.0\% at every NFE \(\geq 8\), and
with \(n=4\), every seed falls below 10\% at NFE \(\geq 16\).
The same runs collapse on 2,000 additional puzzles generated with a different
seed.
With \(R_{\max}=8\) and \(R_{\max}=16\), the extrapolation does not collapse.
With \(R_{\max}=8\), it is up to 1.8 points below the clamp in the mean over
seeds (at most 3.2 points for one seed, at NFE=16); with \(R_{\max}=16\), where
the two differ only at NFE \(\geq 17\), it is 0.1--0.5 points above the clamp
at NFE=32--128 (97.3, 97.8, and 98.0\% versus 97.2, 97.4, and 97.5\%).

On GSM8K, the clamp changes the accuracy by at most 1.8 points in either
direction at NFE=8, 16, and 32 (Table~\ref{tab:gsm8k_clamp}): without the
clamp, \(K=1\) is 0.4--1.8 points higher, \(K=4\) is 0.3--1.5 points lower, and
\(K=12\) is identical at NFE=8 and 16 and 0.7 points higher at NFE=32.
Among the other runs with \(K=12\) in Appendix~\ref{app:gsm8k_ablation} (without
the verifier and the queue, the additive formulation, the two embedding
tables, and \(\tau^{+}=0.1\times\) median), evaluated in the same way, none collapses without the clamp, and the
two differ by at most 1.6 points at NFE=8, 16, and 32.
The GSM8K numbers with the clamp at every checkpoint are listed in
Appendix~\ref{app:gsm8k_full}.
Overall, the clamp costs little on GSM8K and prevents the collapse on Sudoku.

\begin{table*}[tp]
\centering
\small
\begin{tabular}{llrrrrr}
\hline
Condition & Round embedding & NFE=8 & 16 & 32 & 64 & 128 \\
\hline
Source & extrapolated & 39.4\(\pm\)40.2 & 46.4\(\pm\)41.4 & 54.4\(\pm\)47.2 & 57.1\(\pm\)49.5 & 58.0\(\pm\)50.2 \\
 & clamped & 88.4\(\pm\)0.4 & 92.1\(\pm\)0.4 & 93.1\(\pm\)0.6 & 93.5\(\pm\)0.7 & 93.9\(\pm\)0.7 \\
Additive & extrapolated & 38.6\(\pm\)40.3 & 33.5\(\pm\)19.0 & 49.0\(\pm\)26.8 & 55.6\(\pm\)29.6 & 58.4\(\pm\)32.3 \\
 & clamped & 86.2\(\pm\)3.6 & 89.5\(\pm\)3.1 & 90.5\(\pm\)3.0 & 91.0\(\pm\)2.8 & 91.6\(\pm\)2.5 \\
Conditional & extrapolated & 74.7\(\pm\)18.5 & 87.6\(\pm\)1.5 & 89.5\(\pm\)2.5 & 90.1\(\pm\)2.7 & 90.9\(\pm\)3.4 \\
 & clamped & 86.1\(\pm\)1.2 & 89.6\(\pm\)1.8 & 90.7\(\pm\)1.7 & 91.2\(\pm\)2.0 & 91.6\(\pm\)2.3 \\
Source, \(n=4\) & extrapolated & 11.8\(\pm\)20.4 & 1.2\(\pm\)0.9 & 1.8\(\pm\)1.7 & 2.6\(\pm\)2.6 & 2.8\(\pm\)2.9 \\
 & clamped & 87.4\(\pm\)1.5 & 91.1\(\pm\)1.2 & 92.1\(\pm\)1.1 & 92.6\(\pm\)1.2 & 92.7\(\pm\)1.2 \\
Source, \((1,1)\), \(n=4\) & extrapolated & 0.0\(\pm\)0.0 & 1.9\(\pm\)2.2 & 2.9\(\pm\)2.3 & 4.0\(\pm\)3.6 & 4.4\(\pm\)3.7 \\
 & clamped & 88.0\(\pm\)1.2 & 93.5\(\pm\)0.6 & 95.9\(\pm\)0.4 & 97.1\(\pm\)0.3 & 98.0\(\pm\)0.2 \\
Source, \(R_{\max}=3\) & extrapolated & 0.0\(\pm\)0.0 & 0.0\(\pm\)0.0 & 0.0\(\pm\)0.0 & 0.0\(\pm\)0.0 & 0.0\(\pm\)0.0 \\
 & clamped & 71.4\(\pm\)2.1 & 72.6\(\pm\)2.0 & 72.9\(\pm\)1.9 & 73.1\(\pm\)1.9 & 73.1\(\pm\)2.0 \\
Source, \(R_{\max}=8\) & extrapolated & 91.2\(\pm\)1.7 & 93.1\(\pm\)1.9 & 95.5\(\pm\)1.0 & 96.3\(\pm\)0.9 & 96.7\(\pm\)0.7 \\
 & clamped & 91.2\(\pm\)1.7 & 94.9\(\pm\)1.3 & 96.0\(\pm\)1.0 & 96.6\(\pm\)1.0 & 96.8\(\pm\)0.9 \\
Source, \(R_{\max}=16\) & extrapolated & 93.0\(\pm\)1.7 & 96.1\(\pm\)0.8 & 97.3\(\pm\)0.6 & 97.8\(\pm\)0.6 & 98.0\(\pm\)0.6 \\
 & clamped & 93.0\(\pm\)1.7 & 96.1\(\pm\)0.8 & 97.2\(\pm\)0.5 & 97.4\(\pm\)0.4 & 97.5\(\pm\)0.4 \\
\hline
\end{tabular}
\caption{Sudoku Hard accuracy (\%) at 20,000 updates with the round embedding extrapolated or clamped for NFE \(\geq 8\) (same checkpoints; mean \(\pm\) standard deviation over three seeds). For \(R_{\max}=8\) and \(R_{\max}=16\), the two coincide up to NFE=8 and NFE=16, respectively.}
\label{tab:sudoku_clamp}
\end{table*}

\begin{table}[htbp]
\centering
\small
\begin{tabular}{llrrr}
\hline
 & Round embedding & NFE=8 & 16 & 32 \\
\hline
\(K=1\) & extrapolated & 11.1 & 13.3 & 14.7 \\
 & clamped & 10.7 & 12.2 & 12.9 \\
\(K=4\) & extrapolated & 24.0 & 23.4 & 24.9 \\
 & clamped & 24.4 & 24.9 & 25.2 \\
\(K=12\) & extrapolated & 29.0 & 29.4 & 30.1 \\
 & clamped & 29.0 & 29.4 & 29.4 \\
\hline
\end{tabular}
\caption{GSM8K test accuracy (\%) of ReLaDrift at 250,000 updates with the round embedding extrapolated or clamped for NFE \(\geq 8\) (EMA, single; same checkpoints and noise).}
\label{tab:gsm8k_clamp}
\end{table}

\subsection{GSM8K Ablations and Training Progress}\label{app:gsm8k_ablation}

Table~\ref{tab:gsm8k_ablation} ablates ReLaDrift (\(K=12\)) at 250,000 updates
with the clamped round embedding at every NFE, including the rows of
Table~\ref{tab:gsm8k_main_ablation}: the verifier and the positive queue, the
drift-field coefficients, the bandwidth of the attraction, the embedding space,
the formulation, the training depth, and the number of given positives \(K\).
Without the verifier and the positive queue, the accuracy at NFE=16 drops from
29.4\% to 20.0\%; with \((\lambda^{+},\lambda^{-})=(1,1)\), training collapses
to 0.0\% at every NFE by the first evaluation at 20,000 updates, with \(n=2\)
and with \(n=4\), and both runs were stopped at 40,000 updates
(Appendix~\ref{app:lambda}).
The bandwidth of the attraction, \(\tau^{+}=0.01\) times the median distance
between two distinct positives of the same problem
(Appendix~\ref{app:drift_field}), is ablated with the factors 0.1 and 1.
With 0.1, the accuracy is lower at every NFE (10.3\% at NFE=1, 24.0\% at
NFE=4, and 26.3\% at NFE=16, versus 13.2, 27.3, and 29.4\%) but still
increases with NFE, and it is below the default at 12 of the 13 checkpoints
(26.0\% versus 25.9\% at 180,000 updates; Table~\ref{tab:gsm8k_matched}).
With 1, the model, trained for 40,000 updates, reaches at most 0.4\% at every
NFE at 20,000 and 40,000 updates: with such a bandwidth, the attractive weights
no longer concentrate on the nearest positive, and the attraction points toward
a barycenter of several programs, as with the joint normalization discussed in
Appendix~\ref{app:drift_field}.
Replacing the contextualized CodeT5 embeddings with a context-free embedding
table (the CodeT5 input embedding of each token, or a random Gaussian table)
lowers the accuracy to 17.7\% and 7.1\% at NFE=16, but the accuracy still
increases with NFE.
The additive formulation, trained in the same way from scratch, is within
about one point of the source formulation at every NFE of
Table~\ref{tab:gsm8k_ablation} (0.4--1.1 points below at NFE=4--32 and 0.8
points above at NFE=1; 28.4\% versus 29.4\% at NFE=16), and its accuracy also
increases monotonically with NFE up to NFE=32.
Increasing the training depth to \(R_{\max}=8\) or 16 (round embedding clamped
to \(\phi(\min(r,R_{\max}-1))\)) lowers the accuracy at every NFE: by 3.0 and
5.2 points at NFE=1, 3.1 and 3.3 at NFE=4, 1.7 and 2.2 at NFE=16, and 1.2--1.7
at NFE \(\geq 32\), the opposite of the trend on Sudoku
(Table~\ref{tab:sudoku_ablations}).
With the terminal round sampled uniformly, each round receives the loss for a
fraction \(1/R_{\max}\) of the problems, so a larger depth trains each round,
including the first, on fewer problems for the same number of updates.

The compared methods were also evaluated at sampling temperature \(T=1\)
(single, 250,000 updates; NFE=32 for S-FLM, Duo, and MDLM, and sequential
generation for AR): AR 66.0, S-FLM 12.6, Duo 15.0, and MDLM 7.7, versus 72.3,
19.9, 30.6, and 18.8 at \(T=0.1\); the main table therefore reports \(T=0.1\).
Table~\ref{tab:gsm8k_matched} shows the accuracy over training with the
generation settings of the compared methods (AR sequential and S-FLM with 32
steps, both at \(T=1\); FMLM+ with 32 rounds and argmax) and ReLaDrift at
NFE=16 with the clamp.

\begin{table*}[tp]
\centering
\small
\resizebox{\textwidth}{!}{%
\begin{tabular}{lrrrrr|rrrrrr}
\hline
Setting & NFE=1 & 4 & 5 & 8 & 16 & 32 & 64 & 128 & 256 & 512 & 1024 \\
\hline
\((2,1)\), verifier, \(K=12\) (default) & 13.2 & 27.3 & 27.4 & 29.0 & 29.4 & 29.4 & 29.4 & 29.5 & 29.5 & 29.4 & 29.4 \\
without verifier and queue & 4.5 & 17.0 & 17.3 & 18.5 & 20.0 & 20.3 & 20.8 & 20.7 & 20.7 & 20.6 & 20.7 \\
\((1,1)\), verifier (stopped at 40,000 updates) & 0.0 & 0.0 & 0.0 & 0.0 & 0.0 & 0.0 & 0.0 & -- & -- & -- & -- \\
\(\tau^{+}=0.1\times\) median & 10.3 & 24.0 & 24.5 & 24.6 & 26.3 & 26.8 & 26.8 & 27.0 & 27.0 & 27.0 & 27.0 \\
\(\tau^{+}=1\times\) median (40,000 updates) & 0.0 & 0.2 & 0.2 & 0.2 & 0.4 & 0.4 & 0.4 & -- & -- & -- & -- \\
CodeT5 embedding table & 6.7 & 15.1 & 16.0 & 16.7 & 17.7 & 18.0 & 18.2 & 18.1 & 18.1 & 18.1 & 18.2 \\
random embedding table & 1.4 & 5.5 & 6.0 & 6.7 & 7.1 & 7.5 & 7.4 & 7.5 & 7.4 & 7.4 & 7.5 \\
additive formulation & 14.0 & 26.2 & 27.0 & 27.9 & 28.4 & 28.8 & 29.1 & 29.3 & 29.6 & 29.6 & 29.5 \\
single-round Drifting Model & 8.4 & 7.8 & 7.7 & 7.7 & 7.7 & 7.7 & 7.7 & 7.7 & 7.7 & 7.7 & 7.7 \\
architecture of ReLaDrift, one round (control) & 7.9 & -- & -- & -- & -- & -- & -- & -- & -- & -- & -- \\
training depth \(R_{\max}=8\) & 10.2 & 24.2 & 25.0 & 26.2 & 27.7 & 28.1 & 28.2 & 28.2 & 28.1 & 28.2 & 28.1 \\
training depth \(R_{\max}=16\) & 8.0 & 24.0 & 24.9 & 26.8 & 27.2 & 27.7 & 28.0 & 28.2 & 28.2 & 28.1 & 28.1 \\
\(K=1\) & 1.0 & 8.7 & 9.3 & 10.7 & 12.2 & 12.9 & 13.3 & 13.6 & 13.6 & 13.7 & 13.7 \\
\(K=4\) & 10.4 & 22.7 & 23.0 & 24.4 & 24.9 & 25.2 & 25.4 & 25.5 & 25.5 & 25.5 & 25.5 \\
\hline
\end{tabular}}
\caption{GSM8K test ablations at 250,000 updates (\%; EMA, single; round embedding clamped to \(\phi(\min(r,R_{\max}-1))\); \(R_{\max}=5\) unless noted). Unless noted, the source formulation with \(K=12\), \((\lambda^{+},\lambda^{-})=(2,1)\), the attraction bandwidth \(\tau^{+}=0.01\times\) the median distance between distinct positives (Appendix~\ref{app:drift_field}), the verifier, and the positive queue. The single-round Drifting Model (trained in the same way) is iterated at NFE \(\geq 2\), with its decoded and re-encoded program fed back in place of the noise (the dashed line of Figure~\ref{fig:teaser}; Appendix~\ref{app:naive_recurrence}). The \((1,1)\) runs (\(n=2\) and \(n=4\)) were stopped at 40,000 updates after collapsing to 0.0\% at every NFE \(\leq 64\) at 20,000 and 40,000 updates. The \(\tau^{+}=1\times\) median run was trained for 40,000 updates only; its row gives the values at 40,000 updates (0.0\% at every NFE at 20,000 updates). The control row keeps the round embedding and all projections of ReLaDrift but is trained with \(R_{\max}=1\); its refinement projections are never trained, so only NFE=1 is reported. The columns to the right of the vertical line (NFE \(\geq 32\)) lie outside the few-step regime.}
\label{tab:gsm8k_ablation}
\end{table*}

\begin{table*}[tp]
\centering
\small
\begin{tabular}{rrrrrrrr}
\hline
 & & & & \multicolumn{4}{c}{ReLaDrift (\(K=12\)), NFE=16} \\
Updates & AR & S-FLM & FMLM+ & default & w/o ver. & \((1,1)\) & \(\tau^{+}=0.1\) \\
\hline
20k & 48.1 & 0.7 & 3.1 & 9.1 & 6.7 & 0.0 & 5.5 \\
40k & 56.7 & 2.4 & 6.7 & 16.1 & 11.9 & 0.0 & 12.7 \\
60k & 59.8 & 4.7 & 9.0 & 19.6 & 13.9 & -- & 17.2 \\
80k & 62.5 & 5.9 & 11.2 & 22.1 & 16.6 & -- & 20.4 \\
100k & 62.5 & 8.0 & 16.9 & 22.8 & 17.6 & -- & 20.6 \\
120k & 63.5 & 8.9 & 18.0 & 25.2 & 17.7 & -- & 23.1 \\
140k & 62.5 & 10.0 & 18.3 & 25.0 & 18.0 & -- & 24.7 \\
160k & 63.8 & 12.6 & 20.8 & 26.5 & 18.8 & -- & 25.7 \\
180k & 65.2 & 11.4 & 19.5 & 25.9 & 19.3 & -- & 26.0 \\
200k & 66.0 & 13.2 & 21.3 & 26.8 & 18.7 & -- & 26.2 \\
220k & 65.1 & 13.5 & 23.7 & 27.2 & 20.5 & -- & 26.7 \\
240k & 64.6 & 13.2 & 24.7 & 28.7 & 19.9 & -- & 26.8 \\
250k & 66.0 & 12.6 & 24.9 & 29.4 & 20.0 & -- & 26.3 \\
\hline
\end{tabular}
\caption{GSM8K test accuracy (\%) over training (EMA, single). AR and S-FLM sample with temperature \(T=1\) (AR sequential, S-FLM 32 steps), and FMLM+ uses 32 rounds with argmax; ReLaDrift (\(K=12\)) is at NFE=16 with the clamp. ``w/o ver.'' is \((\lambda^{+},\lambda^{-})=(2,1)\) without the verifier and the queue, ``\((1,1)\)'' is \(V^{+}-V^{-}\) with the verifier, stopped at 40,000 updates, and ``\(\tau^{+}=0.1\)'' uses \(\tau^{+}=0.1\times\) median instead of \(0.01\times\) median; the progress of \(K=1\) and \(K=4\) is in Table~\ref{tab:gsm8k_full}.}
\label{tab:gsm8k_matched}
\end{table*}

\subsection{Iterating the Single-Round Drifting Model at Test Time}\label{app:naive_recurrence}

To test whether additional rounds help a Drifting Model that is not trained for
them, we iterate the single-round Drifting Model of each task at test
time.
The first round (NFE=1) is its usual generation from the evaluation noise.
Each later round decodes the output of the previous round, encodes the decoded
sequence as the training targets are encoded (on Sudoku, the one-hot code of
the 89 answer tokens; on GSM8K, the CodeT5 encoding of the prompt, the program
up to the first EOS, and EOS; Appendix~\ref{app:encoding}), and feeds it to the
model in place of the noise; no new noise is drawn.
The noise is the only input through which this model can receive its previous
output, because neither single-round Drifting Model has a round embedding, a draft
projection, or an output projection for the refinement rounds.
A control with the architecture of ReLaDrift, including the round embedding and
all projections, but trained with \(R_{\max}=1\) reaches 7.9\% at NFE=1 on GSM8K
(Table~\ref{tab:gsm8k_ablation}), so the absence of the round embedding does not
explain the gap to ReLaDrift at NFE=1; its refinement projections are never
trained, so it is not iterated.
The evaluation noise, batches, and scoring are those of
Tables~\ref{tab:sudoku_full} and~\ref{tab:gsm8k_ablation}, and NFE=1 reproduces
the evaluation of the single-round Drifting Model for every puzzle and problem.

On Sudoku, the accuracy is 4.9--5.0\% at every NFE from 2 to 128, against
5.1\% at NFE=1 (Table~\ref{tab:sudoku_full}), and the accuracy per cell stays
at 87.9\% at every NFE; from NFE=8 on, about 1,800 of the 2,000 puzzles no
longer change from one round to the next.
On GSM8K, the accuracy is 7.6--7.8\% at every NFE from 2 to 1,024, against
8.4\% at NFE=1 (Table~\ref{tab:gsm8k_ablation}).
The programs are rewritten (at NFE=2, only 249 of the 1,319 programs keep the
code of NFE=1 apart from the docstring), but between NFE=1 and NFE=16, 18
problems become correct and 27 become wrong.
Running more rounds of a model that is not trained to correct its own drafts
therefore does not improve it.

\subsection{All GSM8K Results of ReLaDrift}\label{app:gsm8k_full}

Table~\ref{tab:gsm8k_full} lists the GSM8K test accuracy of ReLaDrift at every evaluated checkpoint and NFE with the clamped round embedding \(\phi(\min(r,4))\) (Appendix~\ref{app:clamp}); for NFE \(\leq 5\), the clamp does not change the round embedding.

\begin{table}[htbp]
\centering
\scriptsize
\setlength{\tabcolsep}{4pt}
\begin{tabular}{llrrrrrrrr}
\hline
\(K\) & Updates & NFE=1 & 2 & 4 & 5 & 8 & 16 & 32 & 64 \\
\hline
1 & 20k & 0.0 & 0.0 & 0.0 & 0.0 & 0.0 & 0.0 & 0.0 & 0.0 \\
 & 40k & 0.0 & 0.3 & 0.2 & 0.3 & 0.4 & 0.4 & 0.5 & 0.5 \\
 & 60k & 0.2 & 0.3 & 0.5 & 0.6 & 0.8 & 1.1 & 1.1 & 1.1 \\
 & 80k & 0.3 & 0.4 & 1.0 & 1.0 & 1.1 & 1.8 & 2.1 & 2.4 \\
 & 100k & 0.3 & 0.5 & 1.3 & 1.4 & 2.4 & 3.5 & 4.5 & 5.2 \\
 & 120k & 0.3 & 0.9 & 2.4 & 3.1 & 3.9 & 5.1 & 6.1 & 6.5 \\
 & 140k & 0.4 & 1.0 & 4.2 & 4.5 & 5.1 & 6.7 & 7.0 & 7.3 \\
 & 160k & 0.4 & 1.8 & 5.5 & 6.0 & 7.0 & 7.7 & 8.3 & 8.6 \\
 & 180k & 0.5 & 2.0 & 6.4 & 6.9 & 8.1 & 9.7 & 10.8 & 11.1 \\
 & 200k & 0.5 & 2.7 & 6.9 & 7.7 & 8.9 & 9.6 & 10.2 & 10.9 \\
 & 220k & 0.8 & 3.8 & 8.0 & 8.6 & 9.9 & 10.6 & 11.5 & 12.2 \\
 & 240k & 0.9 & 4.5 & 8.3 & 9.7 & 10.8 & 12.1 & 12.7 & 13.1 \\
 & 250k & 1.0 & 3.9 & 8.7 & 9.3 & 10.7 & 12.2 & 12.9 & 13.3 \\
\hline
4 & 20k & 0.2 & 0.7 & 2.1 & 2.5 & 3.1 & 3.3 & 3.4 & 3.3 \\
 & 40k & 0.5 & 2.3 & 6.9 & 7.7 & 9.0 & 10.5 & 11.0 & 11.2 \\
 & 60k & 1.9 & 5.8 & 10.7 & 11.8 & 12.7 & 14.0 & 14.5 & 14.4 \\
 & 80k & 2.7 & 8.4 & 13.1 & 14.4 & 15.2 & 15.9 & 16.5 & 16.3 \\
 & 100k & 3.7 & 9.9 & 14.4 & 14.6 & 16.3 & 17.6 & 18.0 & 18.1 \\
 & 120k & 5.8 & 11.7 & 16.4 & 17.8 & 17.8 & 19.0 & 19.3 & 19.6 \\
 & 140k & 6.6 & 13.6 & 17.5 & 18.7 & 18.7 & 20.8 & 21.1 & 21.5 \\
 & 160k & 7.9 & 14.9 & 18.7 & 19.5 & 20.3 & 20.9 & 21.0 & 21.2 \\
 & 180k & 8.6 & 16.1 & 19.9 & 20.5 & 21.6 & 22.2 & 21.8 & 22.4 \\
 & 200k & 8.8 & 16.5 & 20.9 & 21.7 & 21.6 & 22.3 & 22.4 & 22.6 \\
 & 220k & 8.6 & 16.7 & 20.9 & 21.6 & 22.4 & 23.6 & 23.8 & 24.0 \\
 & 240k & 9.9 & 17.7 & 23.0 & 23.0 & 23.2 & 23.8 & 24.1 & 24.3 \\
 & 250k & 10.4 & 18.7 & 22.7 & 23.0 & 24.4 & 24.9 & 25.2 & 25.4 \\
\hline
12 & 20k & 0.5 & 2.0 & 5.6 & 6.3 & 7.7 & 9.1 & 9.7 & 10.0 \\
 & 40k & 2.4 & 7.4 & 12.3 & 13.1 & 14.3 & 16.1 & 16.3 & 16.2 \\
 & 60k & 4.8 & 11.2 & 15.4 & 17.1 & 18.0 & 19.6 & 19.9 & 19.5 \\
 & 80k & 6.3 & 13.6 & 18.2 & 19.1 & 20.5 & 22.1 & 22.2 & 22.7 \\
 & 100k & 7.4 & 16.5 & 21.0 & 21.5 & 22.4 & 22.8 & 23.9 & 24.3 \\
 & 120k & 8.3 & 16.6 & 22.3 & 22.8 & 24.0 & 25.2 & 25.8 & 25.6 \\
 & 140k & 10.1 & 18.0 & 22.3 & 23.0 & 24.3 & 25.0 & 25.5 & 25.7 \\
 & 160k & 11.5 & 19.3 & 24.9 & 24.9 & 25.5 & 26.5 & 27.0 & 26.8 \\
 & 180k & 11.8 & 19.4 & 23.8 & 25.0 & 25.7 & 25.9 & 26.4 & 26.5 \\
 & 200k & 12.1 & 19.9 & 24.8 & 25.2 & 25.8 & 26.8 & 27.2 & 27.3 \\
 & 220k & 12.8 & 20.7 & 25.1 & 25.5 & 25.9 & 27.2 & 27.4 & 27.1 \\
 & 240k & 12.7 & 22.4 & 26.4 & 26.8 & 27.7 & 28.7 & 29.0 & 29.3 \\
 & 250k & 13.2 & 22.4 & 27.3 & 27.4 & 29.0 & 29.4 & 29.4 & 29.4 \\
\hline
\end{tabular}
\caption{GSM8K test accuracy (\%) of ReLaDrift (Source, \((\lambda^{+},\lambda^{-})=(2,1)\), verifier and positive queue) at every checkpoint and NFE with the clamped round embedding (EMA, single).}
\label{tab:gsm8k_full}
\end{table}

\subsection{Generated Samples}\label{app:samples}

The following programs were generated by ReLaDrift (\(K=12\)) at 250,000 updates on GSM8K test problems (EMA weights, the trajectory of Table~\ref{tab:gsm8k_results}, clamped round embedding).
The docstring, which repeats the problem statement inside the program
(Appendix~\ref{app:qwen_prompt}), is generated like the rest of the program; its long lines are wrapped
here for display.
Over the 1,319 test problems, 221 are wrong at NFE=1 and correct at NFE=16,
and 7 are correct at NFE=1 and wrong at NFE=16.

\begingroup\footnotesize
\begin{verbatim}
# Example 1: correct at NFE=1
def simple_math_problem():
    """
    A robe takes 2 bolts of blue fiber and
    half that much white fiberiber. How many
    bolts in total does it take?
    """
    blue_bolts = 2
    white_bolts = blue_bolts / 2
    total_bolts = blue_bolts + white_bolts
    return total_bolts

# Example 2, NFE=1: wrong (returns 9)
def simple_math_problem():
    """
    There are 6 girls in the park. If there
    are twice the number of boys in the park,
    how many kids are in the park?
    """
    girls = 6
    boys = girls / 2
    totalids = girls + boys
    return totalids

# Example 2, NFE=2: wrong (undefined name)
def simple_math_problem():
    """
    There are 6 girls in the park. If there
    are twice the number of boys in the park,
    how many kids are in the park?
    """
    girls = 6
    boys = girls / 2
    totalids = girls + boys
    return kids

# Example 2, NFE=4: correct
#   (identical at NFE=8 and 16)
def simple_math_problem():
    """
    There are 6 girls in the park. If there
    are twice the number of boys in the park,
    how many kids are in the park?
    """
    girls = 6
    boys = girls * 2
    kids = girls + boys
    return kids
\end{verbatim}
\endgroup

\end{document}